\documentclass[sigconf]{acmart}
\AtBeginDocument{%
  }

\copyrightyear{2026}
\acmYear{2026}
\setcopyright{cc}
\setcctype{by}
\acmConference[KDD '26]{Proceedings of the 32nd ACM SIGKDD Conference on Knowledge Discovery and Data Mining V.2}{August 09--13, 2026}{Jeju Island, Republic of Korea}
\acmBooktitle{Proceedings of the 32nd ACM SIGKDD Conference on Knowledge Discovery and Data Mining V.2 (KDD '26), August 09--13, 2026, Jeju Island, Republic of Korea}
\acmDOI{10.1145/3770855.3817464}
\acmISBN{979-8-4007-2259-2/2026/08}

\usepackage{multirow}
\usepackage{natbib}
\usepackage{bm}
\usepackage{url}
\usepackage{balance}
\usepackage{multirow}
\usepackage[table]{xcolor}
\usepackage{makecell} 
\usepackage{xcolor}
\usepackage{float}
\AtBeginDocument{%
  }
\begin{document}

\title{scTranslation: A Comprehensive Benchmark for Single-Cell \\ Multi-Omics Modality Translation}

\author{Jiabei Cheng\textsuperscript{$\dagger$}}
\affiliation{%
  \institution{Westlake University}
  \city{Hangzhou}
  \state{Zhejiang}
  \country{China}}
\affiliation{%
  \institution{Shanghai Jiao Tong University}
  \city{Shanghai}
  \country{China}}
\email{jiabei_cheng@sjtu.edu.cn}

\author{Jingbo Zhou\textsuperscript{$\dagger$}}
\affiliation{%
  \institution{Zhejiang University}
  \city{Hangzhou}
  \state{Zhejiang}
  \country{China}}
\affiliation{%
  \institution{Westlake University}
  \city{Hangzhou}
  \state{Zhejiang}
  \country{China}}
\email{zhoujingbo@westlake.edu.cn}

\author{Jun Xia}
\affiliation{%
  \institution{The Hong Kong University of Science and Technology (Guangzhou)}
  \city{Guangzhou}
  \state{Guangdong}
  \country{China}}
\email{junxia@hkust-gz.edu.cn}

\author{Changkai Li}
\affiliation{%
  \institution{Xidian University}
  \city{Xian}
  \state{Shaanxi}
  \country{China}}
\email{2975177159@qq.com}

\author{Zhen Lei}
\affiliation{%
  \institution{Institute of Automation, Chinese Academy of Sciences}
  \city{Beijing}
  \country{China}}
\email{zlei@nlpr.ia.ac.cn}

\author{Chang Yu*}
\affiliation{%
  \institution{Westlake University}
  \city{Hangzhou}
  \state{Zhejiang}
  \country{China}}
\email{yuchang@westlake.edu.cn}

\author{Stan Z. Li}
\affiliation{%
  \institution{Westlake University}
  \city{Hangzhou}
  \state{Zhejiang}
  \country{China}}
\email{stan.zq.li@westlake.edu.cn} 
\authornote{Corresponding authors. \\ \textsuperscript{$\dagger$}These authors contributed equally to this work. This work was conducted by Jiabei Cheng during an internship at Westlake University, in collaboration with Jingbo Zhou, the project leader, who formulated the paper's framework.}
\renewcommand{\shortauthors}{Jingbo Zhou et al.}

  \begin{abstract}
Simultaneous measurement of multiple omics modalities in single cells enables researchers to gain a more comprehensive understanding of cellular states and regulatory mechanisms. However, due to high experimental costs, significant noise, and incomplete modality coverage, a variety of computational methods for modality translation have emerged in recent years. Despite the development of translation models, there is still a lack of systematic benchmark evaluation in terms of datasets, evaluation metrics, and influencing factors. To address this, we present scTranslation, a comprehensive benchmark for single-cell multi-omics modality translation tasks. It includes diverse translation datasets, integrates state-of-the-art models, and provides a comprehensive evaluation metrics. In addition, we assess model performance under different scenarios, such as feature selection, feature quality, and few-shot settings. These factors significantly affect model performance but have rarely been systematically studied before. Leveraging this benchmark, we conduct a large-scale study of current methods, report many insightful findings that open up new possibilities for future development. The benchmark is open-sourced to facilitate future research. The code is anonymously released at \url{https://github.com/Bunnybeibei/scTranslation}.

\end{abstract}




\maketitle
\vspace{-0.5em}
\begin{figure}[tbp] 
    \centering          
    \includegraphics[width=0.4\textwidth]{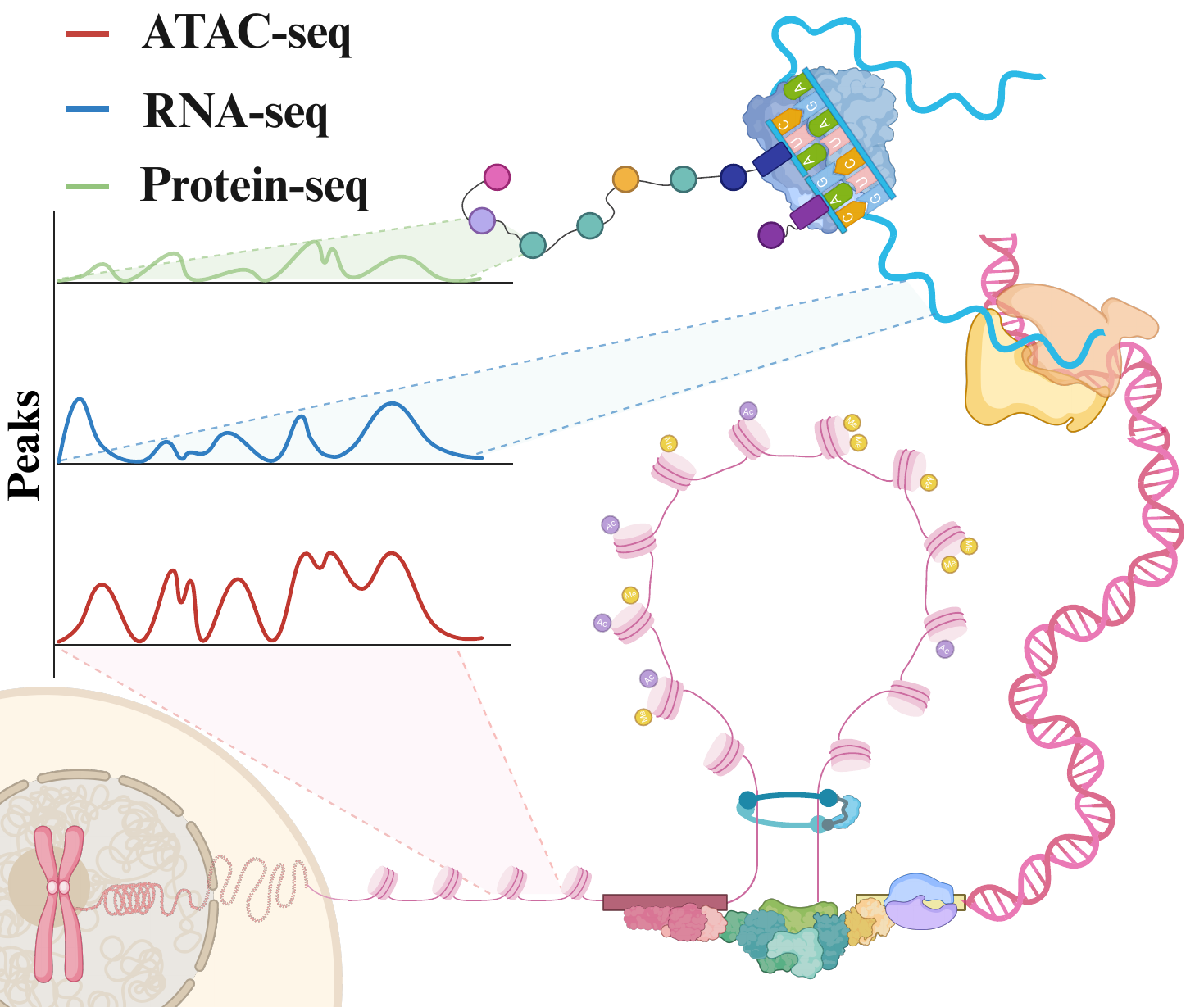}
    \caption{The central dogma of molecular biology describes the flow of genetic information from DNA to RNA to protein. In this schematic, ATAC-seq captures chromatin accessibility, RNA-seq profiles gene expression, and Protein-seq measures protein abundance.} 
    \label{cd}  
    \vspace{-1.5em}
\end{figure}

\section{Introduction}
Over the past decade, single-cell technologies have developed significantly, evolving from measuring a single molecular modality to capturing diverse features such as the transcriptome~\cite{tang2009mrna}, chromatin accessibility, cell surface proteins, and DNA methylation~\cite{Buenrostro2015, Cusanovich2018, clark2018scnmt}. Although protocols such as 10x Multiome, SNARE-seq, and CITE-seq~\cite{stoeckius2017simultaneous} have enabled multi-omic sequencing, they are still constrained by technical complexity, high cost, and severe data sparsity~\cite{kallberg2022frontiers, lim2024advances, subramanian2020multi, bock2016multi}. As a result, their routine application in tasks such as tissue atlas construction~\cite{regev2017human}, large-scale perturbation screens~\cite{dixit2016perturb}, and clinical sample profiling is still limited. Consequently, most publicly available single-cell datasets contain only a single modality, typically RNA or ATAC, leaving critical regulatory or phenotypic information unobserved~\cite{argelaguet2021computational}. To bridge this gap, a new class of computational methods, known as cross-modality translation models, has emerged~\cite{gayoso2021joint, ashuach2023multivi, cao2022multi, gong2021cobolt, argelaguet2020mofa+, argelaguet2018multi, stuart2019comprehensive, athaya2023multimodal, zhang2022semi, lyu2024crossmp}. These models aim to learn statistical mappings between different modalities, enabling tasks such as inferring gene expression programs from chromatin accessibility profiles and reconstructing cis-regulatory structures from transcriptomic data. Cross-modality prediction using deep learning models provides several advantages: \textbf{(i) Cost reduction:} It can reduce sequencing costs substantially compared to true multi-omic experiments by eliminating the need for multiple reagents, deep sequencing, and specialized instrumentation. \textbf{(ii) Denoising:} It mitigates noise in inherently sparse single-cell data. Such data often suffer from dropout, where some genes or peaks are undetected due to low capture efficiency. Cross-modality translation leverages complementary modalities to impute missing signals and reduce noise~\cite{baysoy2023technological, du2022robust}. \textbf{(iii) Augmenting legacy datasets:} It enables multi-omic augmentation of legacy datasets without additional experiments, making historical data comparable to recent multi-omic profiles~\cite{inayatullah2025advances}. \textbf{(iv) Mechanistic insights:} The learned mappings reflect regulatory relationships, such as how chromatin accessibility affects gene expression or how protein abundance influences transcriptional regulation~\cite{lotfollahi2019scgen}.

However, despite their notable successes, existing methods exhibit highly variable performance across different types of datasets and specific tasks. This indicates the necessity of a systematic benchmarking framework to evaluate their accuracy, robustness, and biological fidelity in practical settings~\cite{xiao2024benchmarking, athaya2023multimodal}. Overall, three key challenges remain that limit the future development of deep learning–based single-cell multi-omic modality translation:

\begin{figure}[tbp] 
    \centering          
    \includegraphics[width=0.47\textwidth]{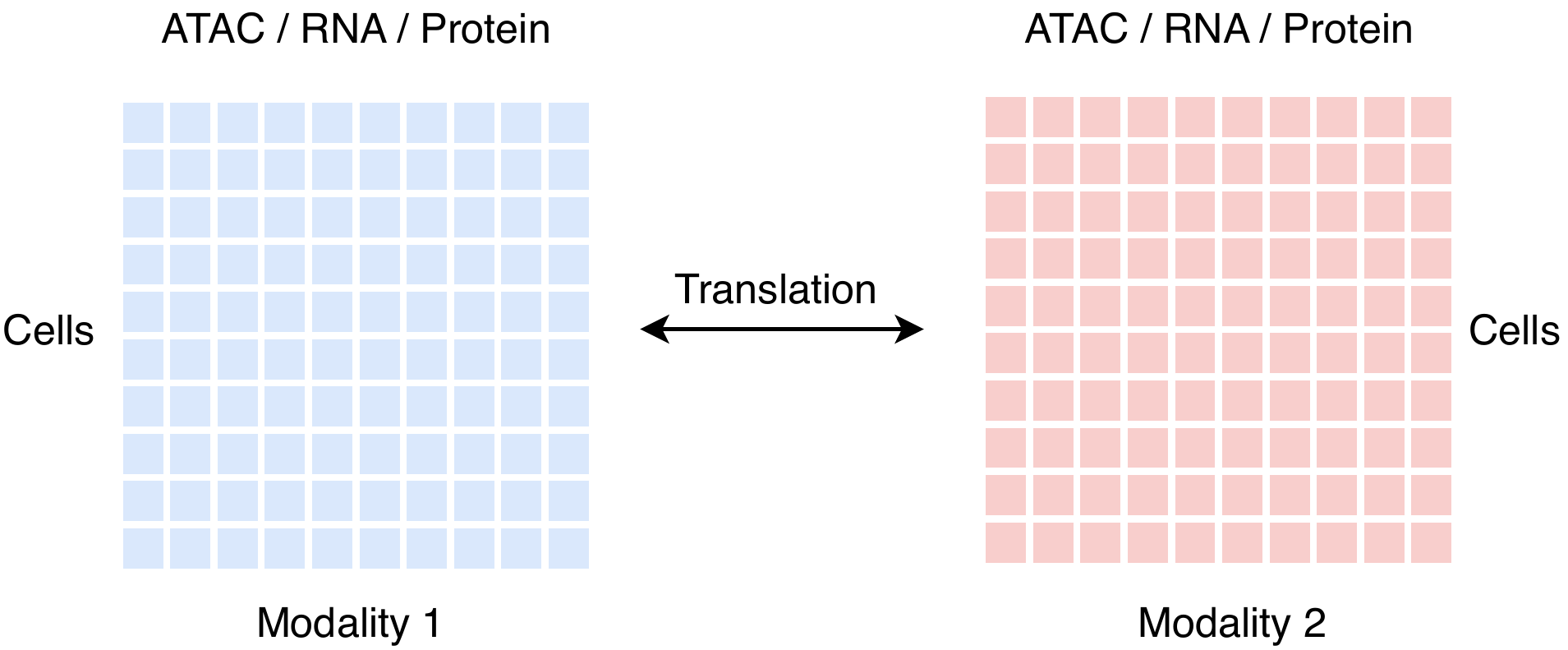}
    \caption{Each modality (e.g., ATAC, RNA, Protein) can be represented as a cell-by-feature matrix. The goal of modality translation is to computationally infer one modality from another, enabling bidirectional prediction between different omics across single cells. } 
    \label{translation_paradigm}  
    \vspace{-1.5em}
\end{figure}

 \begin{itemize}
     \item \textbf{Diverse multi-omic datasets for fair evaluation.} Due to the ease of access to datasets from various single-cell platforms, researchers often use different data sources for training and evaluation, leading to inconsistency across studies. Current methods are typically benchmarked on limited datasets that do not comprehensively cover diverse protocols, species, organs, modalities, scales, or developmental stages. This lack of coverage and standardization may obscure actual progress in the field and hinder fair comparisons between methods.
     \item \textbf{The metrics for comprehensive evaluation.} Evaluating cross-modality translation requires more than a single performance indicator. A multi-faceted evaluation framework is necessary to capture different aspects of model effectiveness, structured across three levels: Clustering-based metrics assess the extent to which the translated data retain cell-type specificity and preserve biological heterogeneity. Regression-based metrics quantify the accuracy of predicted feature values, indicating how well the model captures fine-grained quantitative relationships across modalities. Distribution-level metrics characterize the global statistical alignment between the translated and reference datasets, offering a measure of overall consistency in data distribution.
     \item \textbf{The robustness to important influencing factors.} In practical scenarios, multi-omics data often contains noise, missing values, and incomplete coverage. It is essential for a model to robustly perform under these non-ideal conditions. We conduct evaluations on three key aspects: Varying the feature subset, evaluating whether the model generalizes well across different input configurations, and avoiding dependence on specific preprocessing strategies. Feature corruption and sparsity simulating incomplete or low-quality features to test the model’s resilience to missing data and ability to recover meaningful signals. Few-shot learning assesses the model’s capacity to learn reliable mappings when training data are scarce.
 \end{itemize}

\begin{table*}[tbp]
\centering
\renewcommand{\arraystretch}{1.3}
\caption{Summary of multi-omics translation datasets and their associated metadata.}
\resizebox{\textwidth}{!}{
\renewcommand{\arraystretch}{0.6}
\begin{tabular}{l l l p{3.5cm} p{4.5cm} p{4.5cm}}
\toprule
\textbf{Dataset} & \textbf{Technique} & \textbf{Species} & \textbf{Info} & \textbf{Modality 1} & \textbf{Modality 2} \\
\midrule
GSE126074\_Ad\textbf{Brain}Cortex & SNARE-seq & Adult mouse & Cerebral cortex & RNA: (10309, 31048) & ATAC: (10309, 244544) \\
GSE126074\_\textbf{P0} & SNARE-seq & Neonatal mouse & Cerebral cortex & RNA: (5081, 18257) & ATAC: (5081, 229429) \\
GSM3271041\_\textbf{sciCAR} & sci-CAR & Human & Embryonic kidney & RNA: (4825, 33759) & ATAC: (4825, 189603) \\
\textbf{Brain} & 10x Multiome & Human & Cerebral cortex & RNA: (8981, 34076) & ATAC: (8981, 19241) \\
\textbf{CL} & scCAT-seq & Human & Embryo & RNA: (549, 45434) & ATAC: (549, 157358) \\
\textbf{PBMC} & 10x Multiome & Human & Peripheral blood & RNA: (9631, 28736) & ATAC: (9631, 107194) \\
\textbf{C}ITE\_\textbf{BMMC} & CITE-seq & Human & Bone marrow & RNA: (90261, 13953) & Protein: (90261, 134) \\
\textbf{C}ITE\_\textbf{PBMC} & CITE-seq & Human & Peripheral blood & RNA: (10849, 15230) & Protein: (10849, 14) \\
\bottomrule
\end{tabular}
}
\label{tab:multiomics_datasets}
\end{table*}

\section{Background and Task Definition}
Multi-omic sequencing technologies capture cellular states across regulatory layers such as chromatin regulation, gene transcription, and protein translation. Figure~\ref{cd} shows the flow of genetic information from DNA to RNA to protein. ATAC-seq identifies open chromatin regions through Tn5 transposase-mediated insertion of sequencing adapters~\cite{buenrostro2013transposition}, revealing regulatory elements such as promoters and enhancers. These regions often correspond to transcription factor binding sites and play crucial roles in transcriptional regulation. RNA-seq sequences mRNA transcripts within cells to quantify gene expression levels, providing a direct readout of transcriptional activity. Protein quantification (e.g., via mass spectrometry)~\cite{specht2021single} captures protein abundance, reflecting post-transcriptional regulation such as translation efficiency and protein degradation. Together, these modalities represent a regulatory cascade from chromatin accessibility to protein output: chromatin state influences transcription, which determines mRNA levels and ultimately shapes protein expression. As a result, the relationships among these omics layers are strongly interdependent and nonlinear. However, jointly profiling multiple omics layers at the single-cell level poses substantial technical, economic, and practical challenges. Consequently, large-scale datasets are predominantly limited to single-modality measurements.

Accurate translation of multi-omics modalities enables the integration or imputation of missing data modalities, thereby facilitating downstream analyses such as disease classification, patient stratification, and mechanistic interpretation. Multi-omics modality translation is defined as predicting the molecular profiles of one omics modality (such as transcriptomics, epigenomics, or proteomics) given observed data from another modality. As shown in Figure ~\ref{translation_paradigm}, each input modality is represented as a sparse or dense vector $\mathbf{x} \in \mathbb{R}^M $, where \( M \) is the number of molecular features. Each dimension \( x_i \) corresponds to the observed value \( v_i \) for molecule \( g_i \). The output modality is similarly represented as a vector $\mathbf{y} \in \mathbb{R}^N$, where each dimension \( y_j \) is the predicted value \( \hat{v}_j \) for molecule \( g_j \). Thus, the training dataset consists of two aligned matrices: an input modality matrix $ \mathbf{X} \in \mathbb{R}^{S \times M} $ and an output modality matrix $ \mathbf{Y} \in \mathbb{R}^{S \times N},$ where \( S \) is the number of samples. In single-cell datasets, due to limited sequencing depth, both \( \mathbf{X} \) and \( \mathbf{Y} \) are highly sparse and require preprocessing, such as normalization or feature selection, for stable and generalizable learning.


\section{Datasets}
In this paper, we curate eight datasets to systematically evaluate cross-modality translation methods based on the 6M criteria: multiple techniques, multiple species, multiple organs, multiple omics, multiple scales, and multiple developmental stages. These criteria allow us to assess the stability and generalization ability of existing methods across diverse biological settings. The detailed characteristics of these datasets are summarized in Table~\ref{tab:multiomics_datasets}.

\begin{figure*}[tbp] 
    \centering          
    \includegraphics[width=0.9\textwidth]{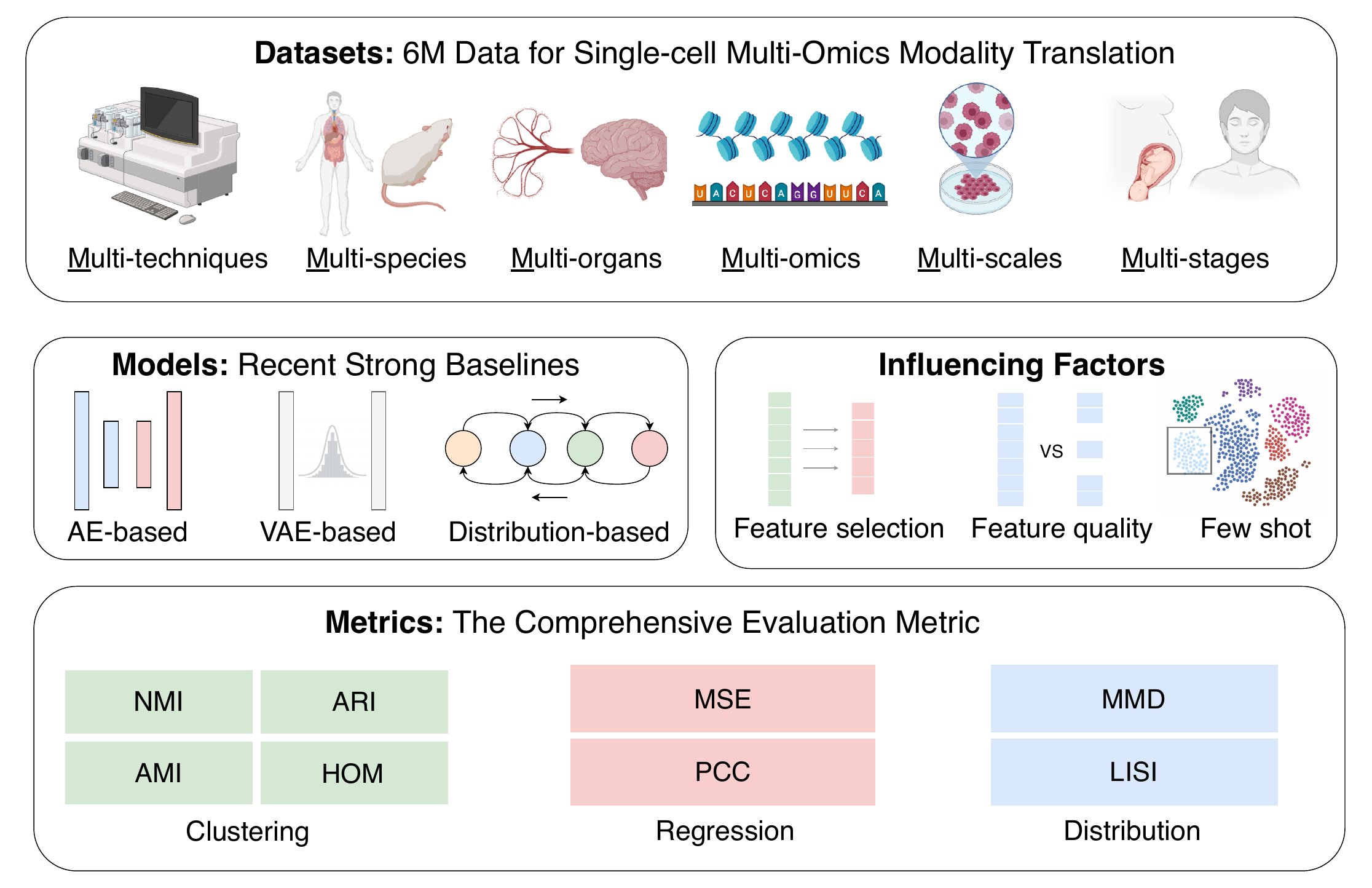}
    \caption{The overview of the scTranslation benchmark. The benchmark is organized incrementally
from datasets to metrics} 
    \label{scTranslation}  
\end{figure*}

\textbf{GSE126074\_AdBrainCortex, GSE126074\_P0.} 
These two datasets are derived from the GSE126074 collection ~\cite{chen2019high}, which contains single-cell multi-omic profiles of the mouse cerebral cortex. The data were generated using SNARE-seq, a protocol that simultaneously profiles RNA and ATAC at the single-nucleus level. SNARE-seq employs Tn5 transposase to tag open chromatin regions while co-capturing mRNA, allowing integration of RNA and ATAC signals under a shared barcode via microfluidic capture. The samples include cerebral cortex tissue from both neonatal mice (postnatal day 0, P0) and adult mice (8 weeks old). The neonatal dataset contains approximately 5,081 nuclei, and the adult dataset contains about 10,309 nuclei. These datasets encompass a variety of neuronal subtypes, glial cells, and a small number of endothelial cells, providing a valuable resource for studying cellular heterogeneity during brain development.

\textbf{GSM3271041\_sciCAR.}
This dataset ~\cite{cao2018joint} contains sci-CAR ATAC-seq data for HEK293T (human embryonic kidney) cells, processed alongside NIH/3T3 and A549 cells. HEK293T cells were cultured at 37°C with 5\% CO$_{2}$ in high-glucose DMEM supplemented with 10\% FBS and 1× Pen/Strep. Both RNA-seq and ATAC-seq reads from HEK293T cells were aligned to a chimeric human-mouse reference genome.

\textbf{Brain.}
This dataset~\cite{trevino2021chromatin} provides a single-cell atlas of gene expression and chromatin accessibility in the developing human cerebral cortex during mid-gestation. Samples were collected from four primary human fetal cortical tissues at post-conception weeks 16, 20, 21, and 24. This dataset was used to identify regulatory waves of key transcription factors along continuous differentiation trajectories, characterize gene expression programs in glial lineages, and uncover transcription factors involved in lineage commitment.

\textbf{CL.}
This dataset~\cite{liu2019deconvolution} was generated using the scCAT-seq protocol with samples derived from clinically discarded human embryos. In total, 110 individual human embryonic cells were profiled. After quality control, 29 cells from the morula stage and 43 cells from the blastocyst stage were retained. This dataset offers insights into dynamic changes in chromatin accessibility and transcriptional activity during early human embryonic development.

\textbf{PBMC.} 
This dataset is a multi-omic peripheral blood mononuclear cell (PBMC) dataset from 10x Genomics, derived from cryopreserved PBMCs of a 25-year-old healthy female donor. Granulocytes were removed by cell sorting. The target recovery was 10,000 nuclei, and approximately 11,898 cells were detected.

\textbf{CITE\_BMMC.}
This dataset~\cite{luecken2021sandbox} contains bone marrow mononuclear cells from 12 healthy human donors. Half of the samples were profiled using the 10x Multiome Gene Expression and Chromatin Accessibility kit, generating paired gene expression and chromatin accessibility data. The remaining samples were processed using the 10x 3' Single-Cell Gene Expression kit combined with the BioLegend TotalSeq B Universal Human Panel v1.0, producing gene expression and surface protein abundance through antibody-derived tags.

\textbf{CITE\_PBMC.}
This dataset is a preprocessed, large-scale single-cell dataset of human peripheral blood mononuclear cells (PBMCs) , jointly profiled for RNA and surface protein expression using CITE-seq from 10x Genomics. It is intended to serve as a high-quality and accessible resource for developing, evaluating, and demonstrating multimodal analysis models within the scVI~\cite{lopez2018deep} ecosystem.

\section{Baseline Models}
Due to recent advances in deep learning, a variety of neural network–based methods have been developed for single-cell multi-omic translation. These methods can be broadly grouped into three categories: Autoencoder (AE)-based models use modality-specific autoencoders and perform cross-modality translation through encoder-decoder pairs. Variational Autoencoder (VAE)-based models adopt probabilistic encoders and decoders to learn latent representations, often using alignment or adversarial objectives to integrate multiple modalities. Distribution-based models leverage a Gaussian mixture model or diffusion processes to perform translation across different omics. We select six recent and representative models—BABEL~\cite{wu2021babel}, JAMIE~\cite{cohen2023joint}, multiDGD~\cite{schuster2024multiDGD}, scButterfly~\cite{cao2024scbutterfly}, scPair~\cite{hu2024scpair}, and scDiffusion-X~\cite{luo2024scdiffusion}—as baseline methods for systematic evaluation.

\textbf{AE-based Models.} AE-based models project single-cell data from different omics into a shared low-dimensional latent space by using modality-specific encoder-decoder architectures~\cite{li2023comprehensive}. This design enables cross-modality reconstruction and translation. BABEL’s architecture consists of four modular neural networks: two encoders and two decoders, each dedicated to either scRNA-seq or scATAC-seq data. The two encoders independently map their input modalities into a shared 16-dimensional latent space, which serves as an integrated representation of cell states. From this latent space, either decoder can reconstruct the corresponding RNA or ATAC profile, enabling flexible translation in both directions. scPair employs paired encoders and decoders for each modality. This design allows the model to learn a shared latent representation that captures the underlying biological state of cells. To ensure the latent space is modality-invariant and biologically meaningful, scPair incorporates adversarial training. A discriminator network attempts to identify the modality source of the latent representations, while the encoder is trained to fool the discriminator, promoting alignment between modalities.

\textbf{VAE-based Models.} VAE-based models use the variational autoencoder~\cite{kingma2013auto} framework to learn modality-specific latent representations and align or combine them in latent space. JAMIE builds on the VAE architecture and addresses the issue of missing modalities by learning separate latent spaces for each omics modality. For samples with partial modality overlap, JAMIE aggregates these latent spaces to form a joint cross-modal representation. To improve interpretability, JAMIE uses Shapley values to identify the most informative input features for cross-modal prediction. scButterfly is based on a dual-aligned VAE framework. It includes two modality-specific encoders, two decoders, a translation module, and two discriminators. Each modality is first pretrained using a masked VAE to extract latent features. These latent spaces are then aligned semantically to capture cross-modal relationships. During training, scButterfly incorporates data augmentation to increase the diversity of training samples, better capture cell-to-cell variability.

\textbf{Distribution-based Models.} Distribution-based models explicitly model the distributional structure of latent variables using parameterized forms such as Gaussian mixtures or diffusion processes. This approach enhances the expressiveness of the latent space, enabling more accurate representation of cellular heterogeneity and cross-modal relationships. multiDGD defines the latent representations as trainable parameters, which improves flexibility and data efficiency. The latent space is modeled using a Gaussian Mixture Model (GMM)~\cite{reynolds2015gaussian}, which is effective in capturing complex cell heterogeneity and feature interactions. Its decoder adopts a hierarchical branching architecture that maps latent variables to outputs corresponding to different omics modalities. multiDGD also incorporates covariates in a probabilistic manner, enabling robust integration across diverse experimental conditions, including previously unseen ones. scDiffusion-X comprises a multimodal autoencoder and a multimodal denoising diffusion network. The autoencoder maps each omics modality into a shared low-dimensional latent space, facilitating efficient integration of heterogeneous data. The denoising diffusion~\cite{ho2020denoising} component iteratively refines representations within this latent space to ensure biologically meaningful reconstruction across modalities. A key innovation of scDiffusion-X is the Dual-Cross-Attention (DCA) module, which provides an adaptive and interpretable mechanism for information exchange between modalities. Furthermore, conditional labels such as cell type, tissue origin, or disease state are embedded into the model, allowing it to generate profiles under specific biological contexts.

\section{Influencing Factors}
In this section, we investigate how various factors influence the performance of cross-modality translation, aiming to evaluate the robustness of models under conditions of feature selection, feature quality, and few-shot learning.

\textbf{Feature Selection.} 
Multi-omic datasets typically involve a large number of genes, but not all genes contribute equally to model training and prediction. Highly variable genes (HVGs), which show substantial expression variation across samples or conditions, often encode more biologically relevant signals and are critical for distinguishing different cellular states or disease phenotypes. Including too many redundant features can increase computational overhead and introduce noise, which may degrade the model’s generalization ability. Selecting representative HVGs allows the model to capture underlying patterns in the data more effectively, improving both the accuracy and efficiency of modality translation. Therefore, evaluating the impact of using different numbers of HVGs is useful for identifying an optimal feature subset that maximizes model performance.

\textbf{Feature Quality.}
Single-cell data are often characterized by high sparsity, with many features being zero in the majority of samples. This sparsity can result from biological factors or technical limitations (such as insufficient sequencing depth). High feature sparsity poses challenges for model training, as it may hinder the learning of robust feature representations and increase the risk of overfitting. Evaluating model performance under varying levels of data sparsity helps assess its robustness to variation in data quality. A robust model should maintain stable performance under moderate sparsity, which is essential for real-world applications, where multi-omic datasets often exhibit substantial variation in completeness and quality.

\textbf{Few-shot Learning.}
In biomedical research, acquiring large-scale, high-quality multi-omic datasets is often expensive and time-consuming. This limitation is especially significant in studies of rare diseases or specific cellular states, where available samples are highly limited. Therefore, few-shot learning ability serves as a critical indicator of model applicability in low-data scenarios. A model with strong few-shot learning capability should be able to learn effective cross-modal mappings and generate accurate translations given only a few training samples. This property is crucial for addressing data scarcity in practical settings, accelerating biological discovery, lowering experimental costs, and broadening the scope of multi-omic technologies.

\begin{table*}[t]
\centering
\caption{Performance comparison (mean $\pm$ standard) of models on multiple evaluation metrics across datasets. The best and the second best are highlighted with \textbf{bold} and \underline{underline}, respectively. MSE and MMD are better when lower, while higher values are better for the other metrics. The table below presents the translation results from \underline{RNA to ATAC} and \underline{RNA to Protein}.}
\label{tab:benchmark-metrics}
\small
\begin{tabular}{ll|cccc|cc|cc}
\toprule
\multirow{2}{*}{\textbf{Dataset}} & \multirow{2}{*}{\textbf{Model}} &
\multicolumn{4}{c|}{\textbf{Clustering}} & \multicolumn{2}{c|}{\textbf{Regression}} & \multicolumn{2}{c}{\textbf{Distribution}} \\
\cmidrule(lr){3-6} \cmidrule(lr){7-8} \cmidrule(lr){9-10}
& & NMI & ARI & AMI & HOM & PCC & MSE & MMD & LISI \\
\midrule
\multirow{6}{*}{\textbf{Brain}} & BABEL & 47.21$_{\pm1.53}$ & 21.68$_{\pm1.65}$ & 45.40$_{\pm1.61}$ & 51.40$_{\pm1.62}$ & \textbf{52.55$_{\pm0.23}$} & 0.1870$_{\pm0.0018}$ & 45.12$_{\pm9.84}$ & \textbf{63.46$_{\pm3.65}$}\\
& scPair & 48.63$_{\pm21.27}$ & 27.31$_{\pm23.33}$ & 46.35$_{\pm22.60}$ & \underline{52.78$_{\pm19.71}$} & 31.62$_{\pm23.93}$ & \textbf{0.0000$_{\pm0.0000}$} & 80.03$_{\pm29.58}$ & 22.66$_{\pm30.70}$\\
& JAMIE & \underline{53.69$_{\pm0.92}$} & \underline{37.58$_{\pm1.92}$} & \underline{52.91$_{\pm0.98}$} & 49.79$_{\pm0.85}$ & 7.44$_{\pm0.26}$ & 1.0331$_{\pm0.0257}$ & \textbf{15.48$_{\pm1.23}$} & 54.26$_{\pm2.62}$\\
& scButterfly & \textbf{66.87$_{\pm1.36}$} & \textbf{41.69$_{\pm3.05}$} & \textbf{65.56$_{\pm1.45}$} & \textbf{74.05$_{\pm1.36}$} & 30.80$_{\pm0.07}$ & 0.0008$_{\pm0.0000}$ & \underline{23.72$_{\pm2.70}$} & \underline{62.12$_{\pm3.89}$}\\
& multiDGD & 3.64$_{\pm5.44}$ & 1.34$_{\pm3.14}$ & 2.39$_{\pm5.73}$ & 3.03$_{\pm4.29}$ & \underline{51.98$_{\pm2.80}$} & \underline{0.0000$_{\pm0.0000}$} & 250.48$_{\pm122.23}$ & 8.96$_{\pm20.03}$\\
& scDiffusion-X & 53.42$_{\pm2.09}$ & 34.31$_{\pm3.05}$ & 52.71$_{\pm2.20}$ & 47.85$_{\pm1.33}$ & 31.53$_{\pm1.34}$ & 0.3271$_{\pm0.0074}$ & 303.10$_{\pm14.67}$ & 0.00$_{\pm0.00}$\\
\midrule
\multirow{6}{*}{\textbf{CL}} & BABEL & \textbf{86.45$_{\pm4.14}$} & \underline{70.93$_{\pm8.64}$} & \textbf{85.41$_{\pm4.47}$} & \textbf{97.04$_{\pm2.86}$} & 15.19$_{\pm1.00}$ & 0.0619$_{\pm0.0032}$ & \underline{60.30$_{\pm4.49}$} & \textbf{83.41$_{\pm9.25}$}\\
& scPair & \underline{83.17$_{\pm1.80}$} & 65.90$_{\pm2.45}$ & \underline{81.83$_{\pm1.92}$} & \underline{95.49$_{\pm2.18}$} & \textbf{55.63$_{\pm1.78}$} & 0.7322$_{\pm0.0863}$ & 119.64$_{\pm9.68}$ & 35.22$_{\pm13.81}$\\
& JAMIE & 79.36$_{\pm5.00}$ & 65.51$_{\pm10.67}$ & 77.94$_{\pm5.38}$ & 86.91$_{\pm5.22}$ & 15.33$_{\pm1.76}$ & 1.0896$_{\pm0.0544}$ & \textbf{37.45$_{\pm5.63}$} & \underline{55.25$_{\pm13.21}$}\\
& scButterfly & 64.78$_{\pm7.05}$ & 47.88$_{\pm7.78}$ & 62.21$_{\pm7.56}$ & 72.69$_{\pm8.06}$ & 10.97$_{\pm2.50}$ & \textbf{0.0113$_{\pm0.0001}$} & 717.93$_{\pm5.73}$ & 0.00$_{\pm0.00}$\\
& multiDGD & 63.64$_{\pm6.86}$ & 42.46$_{\pm7.29}$ & 60.80$_{\pm6.86}$ & 73.03$_{\pm11.05}$ & \underline{55.28$_{\pm1.75}$} & 1.0335$_{\pm0.1023}$ & 236.02$_{\pm13.31}$ & 0.00$_{\pm0.00}$\\
& scDiffusion-X & 74.11$_{\pm12.18}$ & \textbf{71.00$_{\pm12.89}$} & 73.03$_{\pm12.79}$ & 67.70$_{\pm9.24}$ & 42.99$_{\pm1.25}$ & \underline{0.0190$_{\pm0.0026}$} & 118.98$_{\pm10.46}$ & 44.80$_{\pm8.11}$\\
\midrule
\multirow{6}{*}{\textbf{PBMC}} & BABEL & \underline{71.94$_{\pm2.04}$} & \underline{42.93$_{\pm5.76}$} & \underline{71.01$_{\pm2.13}$} & \underline{78.14$_{\pm1.10}$} & 22.39$_{\pm0.12}$ & \underline{0.0240$_{\pm0.0002}$} & \textbf{5.77$_{\pm2.24}$} & \textbf{72.47$_{\pm2.26}$}\\
& scPair & 68.26$_{\pm1.25}$ & 42.41$_{\pm4.73}$ & 67.21$_{\pm1.32}$ & 74.13$_{\pm0.91}$ & \underline{53.10$_{\pm0.13}$} & 0.6766$_{\pm0.0067}$ & 316.47$_{\pm1.46}$ & 0.00$_{\pm0.00}$\\
& JAMIE & 71.80$_{\pm1.05}$ & \textbf{53.51$_{\pm5.40}$} & 70.93$_{\pm1.09}$ & 75.68$_{\pm0.66}$ & 14.60$_{\pm0.14}$ & 0.9841$_{\pm0.0073}$ & \underline{6.28$_{\pm0.20}$} & \underline{70.02$_{\pm1.45}$}\\
& scButterfly & \textbf{74.15$_{\pm3.15}$} & 41.93$_{\pm2.65}$ & \textbf{73.09$_{\pm3.17}$} & \textbf{84.64$_{\pm5.37}$} & 28.41$_{\pm5.47}$ & \textbf{0.0027$_{\pm0.0036}$} & 152.59$_{\pm329.95}$ & 27.91$_{\pm15.92}$\\
& multiDGD & 53.49$_{\pm1.01}$ & 19.56$_{\pm1.56}$ & 51.33$_{\pm1.13}$ & 63.19$_{\pm0.64}$ & \textbf{54.04$_{\pm1.35}$} & 0.8114$_{\pm0.0069}$ & 424.18$_{\pm1.71}$ & 0.00$_{\pm0.00}$\\
& scDiffusion-X & 37.64$_{\pm2.27}$ & 18.51$_{\pm1.95}$ & 35.80$_{\pm2.35}$ & 39.44$_{\pm2.75}$ & 43.09$_{\pm0.26}$ & 0.1182$_{\pm0.0007}$ & 190.85$_{\pm5.69}$ & 0.00$_{\pm0.00}$\\
\midrule
\multirow{6}{*}{\textbf{AdBrainCortex}} & BABEL & \textbf{24.25$_{\pm3.19}$} & \textbf{11.99$_{\pm2.53}$} & \textbf{23.25$_{\pm3.25}$} & \underline{26.05$_{\pm3.21}$} & 8.72$_{\pm0.20}$ & \underline{0.0149$_{\pm0.0002}$} & 11.57$_{\pm2.97}$ & \textbf{63.66$_{\pm2.17}$}\\
& scPair & 13.29$_{\pm1.30}$ & 5.43$_{\pm0.56}$ & 12.38$_{\pm1.26}$ & 15.09$_{\pm1.27}$ & \textbf{20.48$_{\pm0.11}$} & 0.0264$_{\pm0.0005}$ & 16.71$_{\pm1.07}$ & 55.80$_{\pm2.59}$\\
& JAMIE & 21.01$_{\pm2.54}$ & \underline{9.89$_{\pm1.65}$} & 19.79$_{\pm2.50}$ & 22.11$_{\pm2.49}$ & 2.49$_{\pm0.09}$ & 1.0070$_{\pm0.0181}$ & \underline{9.86$_{\pm1.28}$} & \underline{59.30$_{\pm1.68}$}\\
& scButterfly & \underline{22.35$_{\pm1.65}$} & 7.87$_{\pm0.86}$ & \underline{21.11$_{\pm1.79}$} & \textbf{30.06$_{\pm2.40}$} & 11.62$_{\pm0.14}$ & \textbf{0.0004$_{\pm0.0000}$} & \textbf{8.29$_{\pm0.63}$} & 20.41$_{\pm4.44}$\\
& multiDGD & 4.85$_{\pm0.61}$ & 0.38$_{\pm0.13}$ & 1.97$_{\pm0.46}$ & 6.99$_{\pm0.81}$ & \underline{20.42$_{\pm0.12}$} & 0.0283$_{\pm0.0006}$ & 255.01$_{\pm2.75}$ & 0.00$_{\pm0.00}$\\
& scDiffusion-X & 15.90$_{\pm0.91}$ & 9.02$_{\pm1.25}$ & 15.43$_{\pm0.95}$ & 16.32$_{\pm0.66}$ & 13.93$_{\pm0.16}$ & 0.6712$_{\pm0.0067}$ & 309.85$_{\pm16.21}$ & 0.00$_{\pm0.00}$\\
\midrule
\multirow{6}{*}{\textbf{P0}} & BABEL & \underline{18.71$_{\pm0.87}$} & \textbf{9.07$_{\pm1.41}$} & \underline{17.41$_{\pm0.90}$} & \underline{21.27$_{\pm1.06}$} & 10.02$_{\pm0.23}$ & \underline{0.0160$_{\pm0.0002}$} & 16.56$_{\pm1.84}$ & \textbf{59.96$_{\pm2.31}$}\\
& scPair & 14.46$_{\pm2.08}$ & 6.89$_{\pm1.23}$ & 13.18$_{\pm1.89}$ & 16.34$_{\pm1.57}$ & \underline{31.83$_{\pm0.09}$} & 0.0408$_{\pm0.0006}$ & 44.83$_{\pm2.04}$ & 50.65$_{\pm4.08}$\\
& JAMIE & 14.24$_{\pm0.66}$ & 6.95$_{\pm0.87}$ & 12.88$_{\pm0.63}$ & 14.31$_{\pm0.98}$ & 1.24$_{\pm0.07}$ & 1.0548$_{\pm0.0256}$ & \textbf{12.12$_{\pm1.19}$} & \underline{55.50$_{\pm5.74}$}\\
& scButterfly & \textbf{19.60$_{\pm1.05}$} & \underline{7.24$_{\pm1.12}$} & \textbf{17.57$_{\pm1.04}$} & \textbf{26.91$_{\pm1.81}$} & 16.24$_{\pm0.15}$ & \textbf{0.0011$_{\pm0.0001}$} & \underline{13.83$_{\pm1.32}$} & 23.42$_{\pm4.70}$\\
& multiDGD & 7.45$_{\pm0.86}$ & 1.01$_{\pm0.13}$ & 3.95$_{\pm0.57}$ & 10.65$_{\pm0.84}$ & \textbf{32.44$_{\pm0.26}$} & 0.0469$_{\pm0.0008}$ & 314.29$_{\pm3.45}$ & 0.00$_{\pm0.00}$\\
& scDiffusion-X & 8.88$_{\pm0.83}$ & 3.81$_{\pm0.66}$ & 7.90$_{\pm0.86}$ & 10.24$_{\pm1.30}$ & 24.47$_{\pm0.33}$ & 0.7049$_{\pm0.0109}$ & 247.62$_{\pm4.29}$ & 0.00$_{\pm0.00}$\\
\midrule
\multirow{6}{*}{\textbf{sciCAR}} & BABEL & \textbf{41.14$_{\pm0.60}$} & 11.57$_{\pm1.05}$ & \textbf{40.33$_{\pm0.64}$} & \textbf{80.42$_{\pm1.85}$} & 7.80$_{\pm0.30}$ & \underline{0.0123$_{\pm0.0007}$} & 9.96$_{\pm1.17}$ & \textbf{50.64$_{\pm5.16}$}\\
& scPair & \underline{37.74$_{\pm4.13}$} & \underline{13.98$_{\pm3.40}$} & \underline{36.86$_{\pm4.34}$} & 70.96$_{\pm3.35}$ & \underline{21.31$_{\pm4.11}$} & 0.2558$_{\pm0.0341}$ & 114.90$_{\pm5.83}$ & 0.00$_{\pm0.00}$\\
& JAMIE & 32.61$_{\pm2.26}$ & \textbf{14.27$_{\pm2.60}$} & 31.82$_{\pm2.27}$ & 58.44$_{\pm4.81}$ & 8.39$_{\pm0.55}$ & 1.0123$_{\pm0.0785}$ & \textbf{6.99$_{\pm1.33}$} & \underline{16.40$_{\pm4.29}$}\\
& scButterfly & 36.80$_{\pm1.34}$ & 7.26$_{\pm1.02}$ & 35.58$_{\pm1.40}$ & \underline{79.80$_{\pm2.73}$} & 8.33$_{\pm0.25}$ & \textbf{0.0000$_{\pm0.0000}$} & \underline{7.41$_{\pm0.75}$} & 0.31$_{\pm0.24}$\\
& multiDGD & 9.58$_{\pm0.53}$ & 1.17$_{\pm0.15}$ & 7.44$_{\pm0.50}$ & 22.28$_{\pm1.20}$ & \textbf{58.05$_{\pm0.85}$} & 0.2602$_{\pm0.0350}$ & 121.13$_{\pm6.39}$ & 0.00$_{\pm0.00}$\\
& scDiffusion-X & 17.26$_{\pm1.13}$ & 8.72$_{\pm0.66}$ & 16.59$_{\pm1.12}$ & 28.21$_{\pm2.03}$ & 10.46$_{\pm0.19}$ & 525.5168$_{\pm22.0755}$ & 16.12$_{\pm0.26}$ & 0.00$_{\pm0.00}$\\
\midrule
\multirow{6}{*}{\textbf{C\_BMMC}} & BABEL & 36.39$_{\pm2.76}$ & 24.02$_{\pm4.12}$ & 32.87$_{\pm3.16}$ & 34.04$_{\pm2.12}$ & 78.83$_{\pm1.74}$ & 0.0754$_{\pm0.0058}$ & 38.26$_{\pm8.84}$ & 16.04$_{\pm6.75}$\\
& scPair & \underline{77.75$_{\pm0.83}$} & 57.51$_{\pm1.73}$ & \textbf{76.28$_{\pm0.89}$} & \underline{74.01$_{\pm0.80}$} & \textbf{92.98$_{\pm0.09}$} & \textbf{0.0263$_{\pm0.0003}$} & \underline{2.78$_{\pm0.27}$} & \underline{59.47$_{\pm1.99}$}\\
& JAMIE & 77.32$_{\pm0.82}$ & \underline{58.08$_{\pm2.64}$} & 75.94$_{\pm0.85}$ & 72.21$_{\pm1.06}$ & \underline{92.46$_{\pm0.13}$} & \underline{0.0287$_{\pm0.0006}$} & \textbf{1.32$_{\pm0.24}$} & \textbf{74.99$_{\pm2.03}$}\\
& scButterfly & \textbf{77.76$_{\pm0.76}$} & \textbf{58.23$_{\pm3.11}$} & \underline{76.06$_{\pm0.82}$} & \textbf{76.20$_{\pm0.95}$} & 79.06$_{\pm0.07}$ & 0.5634$_{\pm0.0039}$ & 112.18$_{\pm1.13}$ & 0.00$_{\pm0.00}$\\
& multiDGD & 58.37$_{\pm5.76}$ & 29.79$_{\pm2.89}$ & 54.98$_{\pm6.28}$ & 58.72$_{\pm6.00}$ & 79.87$_{\pm6.57}$ & 0.0688$_{\pm0.0200}$ & 60.14$_{\pm27.67}$ & 0.79$_{\pm1.15}$\\
& scDiffusion-X & 25.55$_{\pm2.80}$ & 15.70$_{\pm2.09}$ & 23.53$_{\pm3.16}$ & 20.34$_{\pm1.83}$ & 36.75$_{\pm3.73}$ & 0.3339$_{\pm0.0044}$ & 98.47$_{\pm5.43}$ & 0.00$_{\pm0.00}$\\
\midrule
\multirow{6}{*}{\textbf{C\_PBMC}} & BABEL & 27.91$_{\pm1.67}$ & 16.43$_{\pm1.16}$ & 26.13$_{\pm1.70}$ & 30.52$_{\pm2.04}$ & 72.70$_{\pm1.28}$ & 1.6181$_{\pm0.0645}$ & 43.47$_{\pm3.51}$ & 18.61$_{\pm4.40}$\\
& scPair & \underline{66.49$_{\pm1.09}$} & \underline{39.56$_{\pm1.32}$} & \underline{65.66$_{\pm1.10}$} & \underline{75.39$_{\pm1.54}$} & \textbf{94.14$_{\pm0.24}$} & \textbf{0.3794$_{\pm0.0163}$} & \underline{3.01$_{\pm0.31}$} & \underline{70.81$_{\pm2.94}$}\\
& JAMIE & \textbf{67.39$_{\pm1.52}$} & \textbf{46.53$_{\pm3.84}$} & \textbf{66.70$_{\pm1.56}$} & 73.02$_{\pm1.14}$ & \underline{92.75$_{\pm0.25}$} & \underline{0.4654$_{\pm0.0155}$} & \textbf{3.00$_{\pm0.25}$} & \textbf{80.75$_{\pm1.60}$}\\
& scButterfly & 64.03$_{\pm1.15}$ & 30.78$_{\pm2.82}$ & 62.81$_{\pm1.25}$ & \textbf{77.72$_{\pm0.52}$} & 89.81$_{\pm0.22}$ & 1.2971$_{\pm0.0106}$ & 91.74$_{\pm1.54}$ & 0.00$_{\pm0.00}$\\
& multiDGD & 19.51$_{\pm2.23}$ & 3.70$_{\pm0.63}$ & 15.47$_{\pm1.34}$ & 26.08$_{\pm4.63}$ & 51.66$_{\pm20.11}$ & 2.3063$_{\pm0.5370}$ & 256.16$_{\pm41.08}$ & 0.00$_{\pm0.00}$\\
& scDiffusion-X & 12.40$_{\pm2.85}$ & 8.50$_{\pm2.73}$ & 10.93$_{\pm2.92}$ & 12.87$_{\pm2.92}$ & 59.64$_{\pm1.26}$ & 2.6090$_{\pm0.1156}$ & 70.05$_{\pm9.93}$ & 1.05$_{\pm0.22}$\\
\bottomrule
\end{tabular}
\end{table*}

\section{Metrics}
To comprehensively assess a multi-omics translation model, we divide the metrics into three groups: \emph{clustering consistency}, \emph{regression accuracy}, and \emph{distribution preservation}.  
These three axes respectively interrogate model behaviour on discrete biological structure, continuous expression levels, and global versus local distributional similarity.

\textbf{Clustering Metrics.} Models for single-cell multi-omics translation typically learn to map one modality (e.g., RNA-seq) to another (e.g., ATAC-seq). A successful translation preserves inter-cell differences in the predicted matrix so that true cell types or states remain distinguishable. Accordingly, a family of clustering-consistency metrics tests whether the translated data remain aligned with the real data’s discrete biological structure. \textit{\textbf{NMI (Normalized Mutual Information)}} quantifies the mutual-information overlap between predicted clusters and true labels. It measures shared information: higher values indicate greater retention of cell-type-discriminative signal and remain comparable across experiments or species. \textit{\textbf{ARI (Adjusted Rand Index)}} corrects for chance agreement and evaluates pairwise clustering consistency. In immune datasets or tumor microenvironments, where cell populations are often highly imbalanced, ARI still reveals whether the translation model captures rare subpopulations. \textit{\textbf{AMI (Adjusted Mutual Information)}} further adjusts mutual information for chance, making it suitable for small sample sizes or for severe mismatches in cluster counts; it therefore indicates model reliability under “small-data” or “single-cluster” scenarios. \textit{\textbf{HOM (homogeneity)}} directly measures the internal purity of each predicted cluster. In single-cell translation, a decline in HOM signals that the model over-splits a single cell type or merges distinct types, implying dilution of lineage information.

\textbf{Regression Metrics.} Beyond distinguishing cell types, multi-omics translation must also reproduce the signal intensity of each gene or chromatin peak to support downstream differential analysis and regulatory-network reconstruction. Regression metrics therefore assess quantitative accuracy at the per-gene or per-peak level. \textit{\textbf{PCC (Pearson correlation coefficient)}} measures the linear correlation between predicted and true vectors. A high PCC indicates that the model reproduces overall signal magnitude and preserves coordinated or antagonistic gene-level changes—crucial for inferring co-expression or co-accessibility networks. Conversely, \textit{\textbf{MSE (mean squared error)}} directly quantifies the absolute error at each position. When the translated data serve to detect regulatory-element activity or expression differences, a low MSE ensures that systematic bias does not obscure true signals, whereas a high MSE indicates room for numerical improvement.

\textbf{Distribution Metrics.} A central goal of single-cell multi-omics translation is to integrate predicted and real data into a shared analytical space, enabling joint clustering, trajectory inference, and batch correction. Distribution metrics thus characterize similarity between predicted and real data at both global and local distributional scales. \textit{\textbf{MMD (maximum mean discrepancy)}} assesses global concordance between the two sample distributions. A low MMD suggests the model captures both dominant patterns and rare peaks without mode collapse, whereas a high MMD flags statistically significant deviation from the real distribution. \textbf{\textit{LISI (local inverse Simpson’s index)}} probes each cell’s local neighborhood, quantifying batch mixing (iLISI) and preservation of cell-type boundaries (cLISI). In multi-modal integration, the ideal outcome is higher iLISI with unchanged cLISI, indicating that translation removes technical batch effects without eroding biological heterogeneity.

\vspace{-0.5em}

\section{Results and Analysis}
\subsection{Experimental Settings}
All model hyperparameters are kept identical to those reported in the original papers to ensure a fair comparison. As multiDGD and scDiffusion-X did not originally perform RNA-to-protein translation, we model their initial distributions for this task as multivariate Gaussians. For data preprocessing, we retain the top 3,000 highly variable genes (RNA), remove ATAC peaks present in fewer than 0.5\% of cells, and include all protein channels. Each translation task is evaluated using stratified 5-fold cross-validation, and we report the mean ± standard deviation across folds. All experiments were conducted on a single NVIDIA A100 GPU (80 GB).

\begin{table*}[tbp]
\centering
\caption{Performance comparison (mean $\pm$ standard) of models on multiple evaluation metrics across datasets. The best and the second best are highlighted with \textbf{bold} and \underline{underline}, respectively. MSE and MMD are better when lower, while higher values are better for the other metrics. The table below presents the translation results from \underline{ATAC to RNA} and \underline{Protein to RNA}. BABEL was originally designed for RNA--ATAC translation and does not natively support the protein modality as input; entries marked ``---'' indicate model--task configurations that are not supported, rather than meaningful performance.}
\label{tab:subset_results_r2a}
\small
\begin{tabular}{ll|lccc|cc|cc}
\toprule
\multirow{2}{*}{\textbf{Dataset}} & \multirow{2}{*}{\textbf{Model}} &
\multicolumn{4}{c|}{\textbf{Clustering}} & \multicolumn{2}{c|}{\textbf{Regression}} & \multicolumn{2}{c}{\textbf{Distribution}} \\
\cmidrule(lr){3-6} \cmidrule(lr){7-8} \cmidrule(lr){9-10}
& & NMI & ARI & AMI & HOM & PCC & MSE & MMD & LISI \\
\midrule
\multirow{6}{*}{\textbf{Brain}} & BABEL & 45.13$_{\pm1.47}$ & 20.86$_{\pm2.37}$ & 43.36$_{\pm1.62}$ & 48.35$_{\pm0.84}$ & \underline{93.04$_{\pm1.43}$} & 11.3232$_{\pm4.9143}$ & 46.05$_{\pm25.06}$ & \textbf{66.56$_{\pm12.78}$}\\
& scPair & \underline{54.64$_{\pm2.29}$} & \underline{33.63$_{\pm2.33}$} & \underline{53.46$_{\pm2.30}$} & \underline{56.05$_{\pm3.12}$} & \textbf{93.69$_{\pm0.23}$} & 26.3870$_{\pm1.5995}$ & 263.78$_{\pm9.70}$ & 0.00$_{\pm0.00}$\\
& JAMIE & 51.06$_{\pm1.26}$ & 32.06$_{\pm1.96}$ & 50.09$_{\pm1.34}$ & 48.80$_{\pm0.90}$ & 9.25$_{\pm0.25}$ & \underline{0.9904$_{\pm0.0597}$} & \textbf{9.73$_{\pm3.34}$} & 57.40$_{\pm3.94}$\\
& scButterfly & \textbf{68.18$_{\pm1.09}$} & \textbf{44.01$_{\pm2.28}$} & \textbf{66.98$_{\pm1.18}$} & \textbf{74.89$_{\pm0.90}$} & 71.71$_{\pm0.20}$ & \textbf{0.0510$_{\pm0.0006}$} & \underline{10.95$_{\pm1.46}$} & \underline{62.83$_{\pm1.54}$}\\
& multiDGD & 29.56$_{\pm15.36}$ & 9.54$_{\pm11.34}$ & 25.20$_{\pm16.91}$ & 34.88$_{\pm15.69}$ & 89.46$_{\pm1.34}$ & 26.3811$_{\pm1.5991}$ & 263.69$_{\pm9.69}$ & 0.00$_{\pm0.00}$\\
& scDiffusion-X & 4.29$_{\pm0.47}$ & 1.22$_{\pm0.22}$ & 2.57$_{\pm0.50}$ & 4.08$_{\pm0.44}$ & 84.26$_{\pm1.65}$ & 820.6568$_{\pm67.5727}$ & 66.49$_{\pm21.98}$ & 19.34$_{\pm6.65}$\\
\midrule
\multirow{6}{*}{\textbf{CL}} & BABEL & 78.40$_{\pm2.12}$ & 61.46$_{\pm4.33}$ & 76.60$_{\pm2.38}$ & 90.31$_{\pm3.37}$ & \textbf{64.64$_{\pm1.18}$} & 646.1051$_{\pm120.3272}$ & 107.06$_{\pm21.99}$ & \underline{52.20$_{\pm8.19}$}\\
& scPair & \textbf{87.42$_{\pm1.47}$} & \underline{71.44$_{\pm4.26}$} & \textbf{86.41$_{\pm1.60}$} & \textbf{99.52$_{\pm1.07}$} & \underline{63.78$_{\pm2.27}$} & 1091.3425$_{\pm51.3172}$ & 213.90$_{\pm7.57}$ & 15.19$_{\pm11.49}$\\
& JAMIE & 79.00$_{\pm2.69}$ & 64.02$_{\pm5.57}$ & 77.35$_{\pm3.02}$ & 89.39$_{\pm3.18}$ & 19.86$_{\pm0.65}$ & \underline{1.0517$_{\pm0.0414}$} & \underline{39.29$_{\pm3.38}$} & 34.39$_{\pm10.43}$\\
& scButterfly & \underline{85.66$_{\pm4.09}$} & \textbf{72.42$_{\pm9.83}$} & \underline{84.58$_{\pm4.45}$} & \underline{94.99$_{\pm1.88}$} & 63.14$_{\pm0.91}$ & \textbf{0.6919$_{\pm0.0133}$} & 175.64$_{\pm14.92}$ & 31.04$_{\pm15.24}$\\
& multiDGD & 64.93$_{\pm10.71}$ & 45.67$_{\pm12.36}$ & 62.26$_{\pm11.63}$ & 73.16$_{\pm11.82}$ & 51.80$_{\pm1.49}$ & 1091.3334$_{\pm51.3183}$ & 213.89$_{\pm7.57}$ & 15.17$_{\pm10.97}$\\
& scDiffusion-X & 39.08$_{\pm10.93}$ & 27.28$_{\pm14.75}$ & 34.67$_{\pm12.29}$ & 42.68$_{\pm9.97}$ & 31.09$_{\pm4.14}$ & 795.8074$_{\pm217.1073}$ & \textbf{34.17$_{\pm3.50}$} & \textbf{72.67$_{\pm11.05}$}\\
\midrule
\multirow{6}{*}{\textbf{PBMC}} & BABEL & 59.64$_{\pm5.12}$ & 32.89$_{\pm3.74}$ & 58.37$_{\pm5.26}$ & 64.46$_{\pm5.42}$ & \underline{96.71$_{\pm0.14}$} & 1.7418$_{\pm0.2773}$ & 63.93$_{\pm17.35}$ & 41.52$_{\pm10.94}$\\
& scPair & 63.81$_{\pm2.24}$ & 36.34$_{\pm3.97}$ & 62.71$_{\pm2.34}$ & 68.46$_{\pm2.05}$ & \textbf{96.97$_{\pm0.10}$} & 20.8528$_{\pm3.2180}$ & 423.47$_{\pm26.52}$ & 0.00$_{\pm0.00}$\\
& JAMIE & \textbf{76.77$_{\pm0.87}$} & \textbf{59.51$_{\pm2.11}$} & \textbf{76.16$_{\pm0.86}$} & \textbf{77.36$_{\pm1.04}$} & 22.82$_{\pm0.24}$ & \underline{0.9551$_{\pm0.0307}$} & \textbf{10.16$_{\pm2.93}$} & \textbf{60.73$_{\pm3.11}$}\\
& scButterfly & \underline{66.62$_{\pm18.46}$} & \underline{37.67$_{\pm8.99}$} & \underline{65.44$_{\pm18.74}$} & \underline{75.49$_{\pm23.30}$} & 73.46$_{\pm6.87}$ & \textbf{0.0413$_{\pm0.0092}$} & \underline{49.55$_{\pm86.35}$} & \underline{41.57$_{\pm23.35}$}\\
& multiDGD & 51.92$_{\pm1.71}$ & 19.43$_{\pm1.94}$ & 49.66$_{\pm1.71}$ & 61.41$_{\pm1.94}$ & 90.72$_{\pm2.98}$ & 20.8643$_{\pm3.2197}$ & 423.62$_{\pm26.48}$ & 0.00$_{\pm0.00}$\\
& scDiffusion-X & 50.43$_{\pm2.98}$ & 28.28$_{\pm5.13}$ & 49.16$_{\pm3.16}$ & 51.02$_{\pm2.26}$ & 91.28$_{\pm3.17}$ & 787.6495$_{\pm327.0412}$ & 97.48$_{\pm49.55}$ & 13.27$_{\pm16.04}$\\
\midrule
\multirow{6}{*}{\textbf{AdBrainCortex}} & BABEL & 17.42$_{\pm1.74}$ & 7.32$_{\pm1.16}$ & 16.12$_{\pm1.86}$ & 18.35$_{\pm2.01}$ & \underline{94.82$_{\pm0.32}$} & \underline{0.2532$_{\pm0.0290}$} & 88.25$_{\pm17.29}$ & \textbf{48.45$_{\pm3.75}$}\\
& scPair & 10.56$_{\pm1.46}$ & 3.35$_{\pm0.68}$ & 9.21$_{\pm1.46}$ & 11.36$_{\pm1.44}$ & \textbf{95.40$_{\pm0.18}$} & 3.0127$_{\pm0.1555}$ & 296.99$_{\pm7.27}$ & 0.00$_{\pm0.00}$\\
& JAMIE & \underline{18.60$_{\pm2.51}$} & \underline{9.12$_{\pm1.56}$} & \underline{17.12$_{\pm2.40}$} & \underline{18.83$_{\pm2.67}$} & 5.15$_{\pm0.78}$ & 1.0165$_{\pm0.0339}$ & \textbf{13.27$_{\pm1.92}$} & \underline{46.43$_{\pm2.64}$}\\
& scButterfly & \textbf{37.37$_{\pm1.87}$} & \textbf{18.07$_{\pm1.38}$} & \textbf{36.03$_{\pm1.80}$} & \textbf{44.87$_{\pm3.17}$} & 79.01$_{\pm0.44}$ & \textbf{0.0113$_{\pm0.0003}$} & 133.11$_{\pm8.49}$ & 40.87$_{\pm4.91}$\\
& multiDGD & 5.20$_{\pm0.49}$ & 0.23$_{\pm0.08}$ & 1.26$_{\pm0.47}$ & 6.86$_{\pm0.60}$ & 94.36$_{\pm0.18}$ & 3.0132$_{\pm0.1557}$ & 296.93$_{\pm7.28}$ & 0.00$_{\pm0.00}$\\
& scDiffusion-X & 3.60$_{\pm1.33}$ & 0.56$_{\pm0.51}$ & 2.23$_{\pm1.28}$ & 4.18$_{\pm1.61}$ & 89.64$_{\pm0.94}$ & 1174.5670$_{\pm101.5821}$ & \underline{56.86$_{\pm17.70}$} & 27.29$_{\pm5.14}$\\
\midrule
\multirow{6}{*}{\textbf{P0}} & BABEL & 11.34$_{\pm1.36}$ & 4.32$_{\pm0.57}$ & 9.48$_{\pm1.27}$ & 11.89$_{\pm1.34}$ & \textbf{68.75$_{\pm0.75}$} & \underline{0.0327$_{\pm0.0009}$} & 118.35$_{\pm8.92}$ & \underline{42.82$_{\pm5.32}$}\\
& scPair & \underline{19.19$_{\pm2.55}$} & \textbf{9.79$_{\pm2.09}$} & \textbf{17.66$_{\pm2.70}$} & \underline{21.14$_{\pm2.86}$} & 41.76$_{\pm0.16}$ & 0.1271$_{\pm0.0038}$ & 231.40$_{\pm3.77}$ & 0.00$_{\pm0.00}$\\
& JAMIE & 8.02$_{\pm1.06}$ & 3.54$_{\pm0.82}$ & 6.29$_{\pm1.17}$ & 7.64$_{\pm1.16}$ & 0.61$_{\pm0.10}$ & 1.0419$_{\pm0.0285}$ & \textbf{78.55$_{\pm9.25}$} & \textbf{44.26$_{\pm2.62}$}\\
& scButterfly & \textbf{19.84$_{\pm1.51}$} & \underline{8.12$_{\pm1.22}$} & \underline{16.45$_{\pm1.41}$} & \textbf{23.98$_{\pm0.98}$} & \underline{56.53$_{\pm0.26}$} & \textbf{0.0075$_{\pm0.0001}$} & 181.38$_{\pm2.56}$ & 35.73$_{\pm3.28}$\\
& multiDGD & 8.92$_{\pm0.63}$ & 1.19$_{\pm0.22}$ & 4.72$_{\pm0.75}$ & 12.02$_{\pm0.81}$ & 39.12$_{\pm0.27}$ & 0.1271$_{\pm0.0038}$ & 231.31$_{\pm3.76}$ & 0.00$_{\pm0.00}$\\
& scDiffusion-X & 5.54$_{\pm0.70}$ & 0.92$_{\pm0.65}$ & 2.89$_{\pm0.86}$ & 8.41$_{\pm0.98}$ & 8.89$_{\pm0.72}$ & 545.4109$_{\pm41.4174}$ & \underline{78.83$_{\pm5.51}$} & 0.06$_{\pm0.14}$\\
\midrule
\multirow{6}{*}{\textbf{sciCAR}} & BABEL & 38.48$_{\pm2.14}$ & 13.11$_{\pm2.27}$ & 37.71$_{\pm2.21}$ & 71.63$_{\pm2.05}$ & \underline{30.39$_{\pm0.35}$} & \underline{0.3741$_{\pm0.0245}$} & 45.20$_{\pm4.66}$ & 29.35$_{\pm4.69}$\\
& scPair & \textbf{47.85$_{\pm1.34}$} & \textbf{19.33$_{\pm2.00}$} & \textbf{47.31$_{\pm1.37}$} & \textbf{82.70$_{\pm1.14}$} & 30.13$_{\pm1.07}$ & 0.4679$_{\pm0.0935}$ & 92.11$_{\pm15.75}$ & 0.00$_{\pm0.00}$\\
& JAMIE & \underline{39.42$_{\pm2.85}$} & \underline{17.34$_{\pm3.09}$} & \underline{38.75$_{\pm2.91}$} & 67.13$_{\pm2.14}$ & 9.01$_{\pm0.64}$ & 1.0376$_{\pm0.0551}$ & \underline{31.42$_{\pm4.14}$} & \textbf{39.25$_{\pm8.90}$}\\
& scButterfly & 35.24$_{\pm0.68}$ & 7.92$_{\pm0.37}$ & 34.13$_{\pm0.67}$ & \underline{74.07$_{\pm1.47}$} & \textbf{32.84$_{\pm0.47}$} & \textbf{0.0391$_{\pm0.0005}$} & 102.59$_{\pm4.18}$ & \underline{38.64$_{\pm3.82}$}\\
& multiDGD & 9.55$_{\pm0.60}$ & 1.17$_{\pm0.12}$ & 7.40$_{\pm0.59}$ & 22.23$_{\pm1.38}$ & 29.58$_{\pm0.60}$ & 0.4678$_{\pm0.0935}$ & 91.96$_{\pm15.75}$ & 0.00$_{\pm0.00}$\\
& scDiffusion-X & 16.17$_{\pm2.40}$ & 4.82$_{\pm1.21}$ & 14.74$_{\pm2.45}$ & 31.65$_{\pm4.50}$ & 9.34$_{\pm0.84}$ & 1447.0701$_{\pm44.6684}$ & \textbf{30.29$_{\pm3.41}$} & 14.74$_{\pm8.63}$\\
\midrule
\multirow{6}{*}{\textbf{C\_BMMC}} 
& BABEL        & --- & --- & --- & --- & --- & --- & --- & ---\\
& scPair       & 66.85$_{\pm0.74}$ & 37.92$_{\pm1.91}$ & 64.40$_{\pm0.87}$ & 65.81$_{\pm0.72}$ & \textbf{95.51$_{\pm0.18}$} & 1534004.2683$_{\pm92717.3115}$ & 7.20$_{\pm0.44}$ & 0.00$_{\pm0.00}$\\
& JAMIE        & \underline{77.11$_{\pm1.30}$} & \textbf{58.83$_{\pm2.64}$} & \textbf{75.69$_{\pm1.32}$} & \underline{72.59$_{\pm1.84}$} & 34.63$_{\pm0.31}$ & 1.4001$_{\pm0.1024}$ & \textbf{2.70$_{\pm0.42}$} & \textbf{68.27$_{\pm4.22}$}\\
& scButterfly  & \textbf{77.20$_{\pm1.12}$} & \underline{58.38$_{\pm1.92}$} & \underline{75.58$_{\pm1.19}$} & \textbf{74.61$_{\pm1.04}$} & 77.25$_{\pm0.23}$ & \textbf{0.0510$_{\pm0.0005}$} & \underline{7.20$_{\pm0.44}$} & \underline{44.15$_{\pm4.83}$}\\
& multiDGD     & 54.10$_{\pm5.14}$ & 25.16$_{\pm3.14}$ & 50.11$_{\pm5.49}$ & 55.46$_{\pm5.69}$ & 81.25$_{\pm4.23}$ & 1534008.1095$_{\pm92715.9356}$ & \underline{7.20$_{\pm0.44}$} & 0.00$_{\pm0.00}$\\
& scDiffusion-X & 44.35$_{\pm1.45}$ & 26.73$_{\pm4.56}$ & 40.91$_{\pm1.79}$ & 41.81$_{\pm1.01}$ & \underline{83.49$_{\pm1.64}$} & \underline{4092.0682$_{\pm560.1750}$} & 95.26$_{\pm21.68}$ & 1.23$_{\pm1.33}$\\
\midrule

\multirow{6}{*}{\textbf{C\_PBMC}} 
& BABEL        & --- & --- & --- & --- & --- & --- & --- & ---\\
& scPair       & 60.20$_{\pm0.75}$ & 29.57$_{\pm3.02}$ & 58.98$_{\pm0.83}$ & 71.89$_{\pm0.61}$ & \textbf{95.32$_{\pm0.09}$} & 41.8212$_{\pm4.6781}$ & 277.43$_{\pm15.09}$ & 0.00$_{\pm0.00}$\\
& JAMIE        & \textbf{63.71$_{\pm0.53}$} & \textbf{38.54$_{\pm1.86}$} & \textbf{62.77$_{\pm0.57}$} & \underline{72.58$_{\pm0.45}$} & 23.43$_{\pm0.47}$ & \underline{0.9556$_{\pm0.0150}$} & \underline{17.60$_{\pm1.71}$} & \underline{49.74$_{\pm5.52}$}\\
& scButterfly  & \underline{63.52$_{\pm1.37}$} & \underline{33.97$_{\pm3.76}$} & \underline{62.39$_{\pm1.47}$} & \textbf{75.15$_{\pm1.04}$} & 85.43$_{\pm0.15}$ & \textbf{0.0400$_{\pm0.0003}$} & \textbf{5.80$_{\pm0.20}$} & \textbf{53.59$_{\pm1.70}$}\\
& multiDGD     & 21.23$_{\pm1.72}$ & 4.96$_{\pm3.37}$ & 17.21$_{\pm2.93}$ & 28.00$_{\pm1.79}$ & 70.42$_{\pm28.96}$ & 41.8307$_{\pm4.6758}$ & 277.46$_{\pm15.12}$ & 0.00$_{\pm0.00}$\\
& scDiffusion-X & 45.74$_{\pm2.06}$ & 29.40$_{\pm3.65}$ & 44.76$_{\pm2.13}$ & 48.28$_{\pm1.88}$ & \underline{88.63$_{\pm0.44}$} & 385.2498$_{\pm20.3509}$ & 32.27$_{\pm8.21}$ & 24.20$_{\pm6.20}$\\
\bottomrule
\end{tabular}
\end{table*}

\vspace{-0.5em}

\subsection{Main Results}
\textbf{ATAC \& RNA.}
Across the bidirectional translation results (ATAC \\ $\rightarrow$RNA and RNA$\rightarrow$ATAC), models exhibit clear trade-offs among \emph{clustering}, \emph{regression}, and \emph{distribution} metrics, and these trade-offs are highly dataset-dependent. Overall, VAE-family methods (notably scButterfly and JAMIE) tend to achieve better structural preservation and reconstruction accuracy. In the ATAC$\rightarrow$RNA direction, scButterfly attains top performance on NMI/ARI/AMI/HOM for Brain and AdBrainCortex, and on NMI/HOM for P0 (with scPair leading ARI/AMI), while also yielding the lowest MSE (e.g., Brain: 0.0510, AdBrainCortex: 0.0113, P0: 0.0075), indicating more stable behavior in both cell-type geometry preservation and per-cell reconstruction. On PBMC, JAMIE dominates clustering metrics and distribution distance (MMD), together with higher LISI, suggesting stronger population-level distribution matching. In contrast, AE-based approaches (e.g., BABEL and scPair) often excel in correlation (PCC): for ATAC$\rightarrow$RNA, BABEL/scPair achieve near-best or best PCC on several datasets (e.g., Brain/PBMC/AdBrainCortex), yet their MSE and distribution metrics are not consistently optimal. This implies that high correlation does not necessarily translate to accurate amplitude/scale reconstruction or good distributional alignment. Distribution-oriented methods (e.g., scDiffusion-X) can show marked advantages on distribution metrics for specific datasets (e.g., best LISI and MMD on CL), but may sacrifice clustering structure preservation or incur larger reconstruction error, reflecting an objective that prioritizes global distribution matching. Regarding directionality, BABEL achieves substantially higher PCC in ATAC$\rightarrow$RNA than in RNA$\rightarrow$ATAC, suggesting a stronger decoder for inferring expression from accessibility. Meanwhile, scButterfly shows relatively symmetric clustering performance across directions on several datasets (e.g., Brain and P0 remain top-tier in both directions), although distribution distance (MMD) and mixing (LISI) remain dataset-dependent (e.g., RNA$\rightarrow$ATAC on CL exhibits notably larger MMD), indicating that directional consistency does not necessarily hold simultaneously for all metrics.

\begin{figure*}[tbp]
    \centering
    \includegraphics[width=\textwidth]{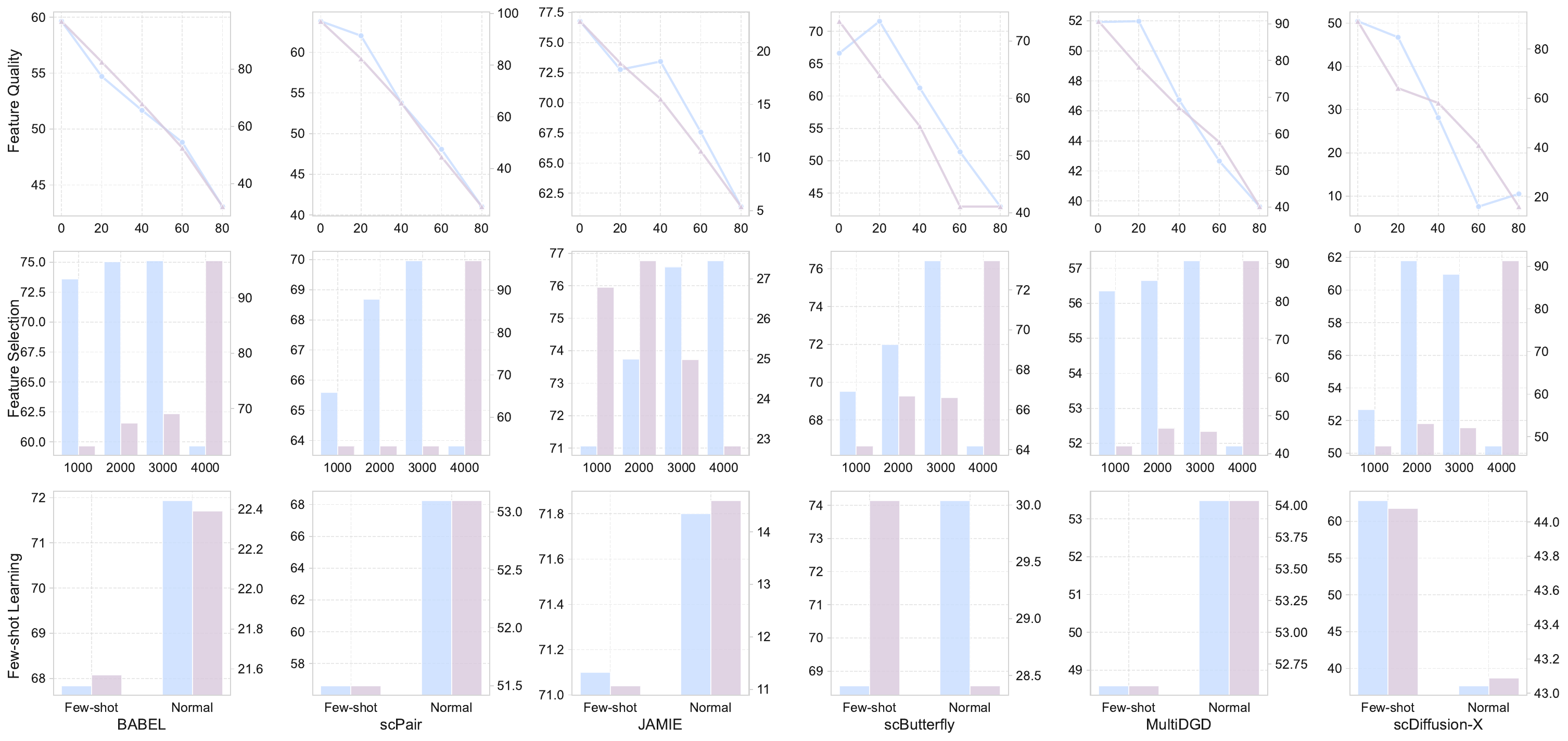}
    \caption{Performance comparison of six translation models under three influencing factors: feature quality (top row), feature selection (middle row), and few-shot learning (bottom row). \underline{Blue} represents the NMI metric, with values corresponding to the left y-axis. \underline{Purple} represents the PCC metric, with values corresponding to the right y-axis. The above figure shows the impact of factors on the performance of RNA to ATAC on the PBMC dataset. 
}
    \label{fig:myfigure}
\vspace{-1em}
\end{figure*}

\textbf{RNA \& Protein.}
On CITE-seq datasets, bidirectional translation highlights a gap between \emph{numerical stability/scale robustness} and \emph{distribution alignment}. In Protein$\rightarrow$RNA, MSE varies dramatically (from near-zero to the $10^6$ scale), implying high sensitivity to normalization, scale, and convergence. VAE-based models are generally more robust: JAMIE and scButterfly stay top-tier on clustering and distribution mixing (JAMIE best LISI on C\_BMMC, scButterfly best LISI on C\_PBMC), and scButterfly attains smaller reconstruction error (C\_BMMC/C\_PBMC MSE: 0.0510/0.0400). By contrast, scPair yields very high PCC (close to 95) on both CITE datasets yet extremely large MSE (e.g., $10^6$-level on C\_BMMC), consistent with ``trend agreement but severe scale mismatch.'' Notably, the protein modality is not natively supported by some models in this direction (e.g., BABEL, whose Protein$\rightarrow$RNA entries are marked ``---''), indicating that model usability and implementation details require particular caution. The RNA$\rightarrow$Protein direction is comparatively more stable: JAMIE performs best on distribution metrics (best MMD/LISI on C\_BMMC and best LISI on C\_PBMC) while maintaining strong clustering consistency, whereas scPair is more competitive on pointwise regression (often best PCC and MSE) but lags behind JAMIE in distribution mixing. Meanwhile, scButterfly degrades substantially on distribution metrics for RNA$\rightarrow$Protein (near-zero LISI and larger MMD on both datasets), revealing pronounced directional asymmetry: it may preserve cell-type structure and reconstruct well in certain directions but struggles with distributional alignment in protein space. Overall, VAE methods (especially JAMIE) offer a more balanced profile in bidirectional distribution alignment and robustness, while AE methods tend to maximize correlation at the risk of scale inconsistency; Protein$\rightarrow$RNA in particular demands careful attention to numerical stability and unsupported configurations.

\subsection{Influencing Factors}
In this section, we evaluate the robustness of all models under varying degrees of three influencing factors: feature selection, feature quality, few-shot Learning.

\textbf{Feature Selection.} In the RNA-to-ATAC translation task, we systematically evaluated the effect of using 500 to 4,000 HVGs as input. As the number of genes increased, distribution metrics exhibited no consistent trend, while clustering metrics initially improved and then declined. Specifically, for up to 3,000 HVGs, the rapidly added information significantly enhanced cell type separability. However, beyond this point, the accumulation of technical noise reduced the signal-to-noise ratio, resulting in decreased clustering consistency. In contrast, regression accuracy improved steadily with increasing input dimensionality, as the inclusion of redundant features expanded the model's capacity to fit the data and more comprehensively captured the underlying variation space.

\textbf{Feature Quality.} When randomly masking 20\%–80\% of expression values to simulate sequencing dropout, we observed a decline in both clustering and regression metrics as the missing rate increased. This decline is attributed to the obscuring of key cell-type–specific signals, an increased proportion of noise, and the model's reduced ability to capture intercellular differences and quantitative relationships. Distribution-based metrics improved at higher dropout levels. This can be explained by the model’s tendency toward mean-averaging behavior under severe missingness—that is, generating outputs that resemble global averages or exhibit reduced variance. While this behavior compromises cell-level specificity, it decreases the MMD distance to the true population-level distribution. These findings suggest that under substantial data loss, models tend to trade off individual resolution for global distributional consistency.

\textbf{Few-shot Learning.} In the few-shot setting, we trained on one fold and evaluated on the remaining four folds. Most models exhibited degraded performance relative to the full-data setting, owing to the drastic reduction in available samples and the consequent imprecise estimation of key feature distributions. Interestingly, scDiffusion-X improved under this setting. This ostensibly counterintuitive result can be attributed to its diffusion-based approach, which naturally provides implicit data augmentation and enables effective exploration of the underlying data manifold even with very limited labeled data. Moreover, the performance gap between the few-shot and normal settings was modest, suggesting that multi-omics data exhibit a degree of redundancy.

\section{Conclusion}
In this work, we present scTranslation, a unified framework for the systematic evaluation of single-cell multi-omics modality translation across data, models, influencing factors, and metrics. We incorporate the 6M data dimension and three major modeling paradigms into an open-source benchmark, substantially lowering barriers to reproducibility and to the development of new methods. Within a fair and reproducible evaluation protocol, we find no universally optimal model. Performance is highly sensitive to the feature-selection scale, feature quality and sparsity, and low-data regimes. These observations underscore the need for future methods to prioritize robustness and scalability. We hope that scTranslation will serve as an important resource to advance the development and practical application of multi-omics modality translation methods.

\section*{Acknowledgments}
This work was supported by the National Science and Technology Innovation - Major Program (No. 2022ZD0115101), the National Science and Technology Major Project of China (No. 2021YFA1301603), the Project (No. WU2025B006) from the SOE Dean Special Project Fund (SOE-DSPF) Program of Westlake University, and the Open Research Fund of the State Key Laboratory of Multimodal Artificial Intelligence Systems (No. MAIS2025064).

\newpage
\bibliographystyle{ACM-Reference-Format}
\bibliography{main}

\clearpage
\appendix

\clearpage

\begin{table*}[htbp]
\centering
\caption{The table below presents the translation results from RNA to ATAC on \underline{\textbf{feature selection}} factor.}
\label{tab:length_results_r2a}
\scriptsize
\renewcommand{\arraystretch}{0.5}
\setlength{\tabcolsep}{4pt}
\begin{tabular}{lll|cccc|cc|cc}
\toprule
\multirow{2}{*}{\textbf{Dataset}} & \multirow{2}{*}{\textbf{Model}} & \multirow{2}{*}{\textbf{Length}} & \multicolumn{4}{c|}{\textbf{Clustering}} & \multicolumn{2}{c|}{\textbf{Regression}} & \multicolumn{2}{c}{\textbf{Distribution}} \\
\cmidrule(lr){4-7} \cmidrule(lr){8-9} \cmidrule(lr){10-11}
 & & & \textbf{NMI} & \textbf{ARI} & \textbf{AMI} & \textbf{HOM} & \textbf{PCC} & \textbf{MSE} & \textbf{MMD} & \textbf{LISI} \\
\midrule
\multirow{30}{*}{Brain} & \multirow{5}{*}{BABEL} & 500 & \underline{48.50}$_{\pm1.63}$ & \underline{24.71}$_{\pm1.18}$ & \underline{46.83}$_{\pm1.63}$ & \underline{52.17}$_{\pm2.01}$ & 52.54$_{\pm0.13}$ & 0.19$_{\pm0.00}$ & 68.24$_{\pm9.73}$ & \underline{55.51}$_{\pm2.56}$ \\
 &  & 1000 & \textbf{51.16}$_{\pm2.17}$ & \textbf{26.20}$_{\pm2.22}$ & \textbf{49.49}$_{\pm2.29}$ & \textbf{55.48}$_{\pm1.98}$ & 52.52$_{\pm0.17}$ & \underline{0.19}$_{\pm0.00}$ & 72.78$_{\pm17.51}$ & \textbf{53.57}$_{\pm5.50}$ \\
 &  & 2000 & 48.24$_{\pm2.85}$ & 23.54$_{\pm1.79}$ & 46.56$_{\pm2.88}$ & 51.77$_{\pm3.27}$ & \underline{52.56}$_{\pm0.22}$ & 0.19$_{\pm0.00}$ & 57.56$_{\pm17.72}$ & 59.69$_{\pm6.54}$ \\
 &  & 3000 & 47.21$_{\pm1.53}$ & 21.68$_{\pm1.65}$ & 45.40$_{\pm1.61}$ & 51.40$_{\pm1.62}$ & 52.55$_{\pm0.23}$ & 0.19$_{\pm0.00}$ & \textbf{45.12}$_{\pm9.84}$ & 63.46$_{\pm3.65}$ \\
 &  & 4000 & 47.42$_{\pm1.90}$ & 21.64$_{\pm2.04}$ & 45.58$_{\pm2.01}$ & 51.76$_{\pm2.00}$ & \textbf{52.65}$_{\pm0.18}$ & \textbf{0.18}$_{\pm0.00}$ & \underline{46.39}$_{\pm12.25}$ & 56.16$_{\pm5.88}$ \\
\cmidrule{2-11}
 & \multirow{5}{*}{scPair} & 500 & 27.51$_{\pm3.26}$ & 8.05$_{\pm1.93}$ & 24.20$_{\pm3.40}$ & 31.38$_{\pm3.79}$ & 5.73$_{\pm7.59}$ & \textbf{0.00}$_{\pm0.00}$ & 147.48$_{\pm41.22}$ & \textbf{1.48}$_{\pm2.43}$ \\
 &  & 1000 & 38.79$_{\pm16.94}$ & 17.61$_{\pm17.37}$ & 35.92$_{\pm18.04}$ & 43.42$_{\pm16.42}$ & 18.09$_{\pm21.88}$ & \underline{0.00}$_{\pm0.00}$ & 114.38$_{\pm54.45}$ & \underline{11.58}$_{\pm25.03}$ \\
 &  & 2000 & 41.61$_{\pm16.05}$ & 19.64$_{\pm16.96}$ & 38.94$_{\pm17.21}$ & 46.00$_{\pm14.88}$ & 16.80$_{\pm27.05}$ & 0.00$_{\pm0.00}$ & 122.63$_{\pm73.97}$ & 16.86$_{\pm24.32}$ \\
 &  & 3000 & \underline{48.63}$_{\pm21.27}$ & \underline{27.31}$_{\pm23.33}$ & \underline{46.35}$_{\pm22.60}$ & \textbf{52.78}$_{\pm19.71}$ & \underline{31.62}$_{\pm23.93}$ & 0.00$_{\pm0.00}$ & \underline{80.03}$_{\pm29.58}$ & 22.66$_{\pm30.70}$ \\
 &  & 4000 & \textbf{49.30}$_{\pm18.54}$ & \textbf{27.80}$_{\pm19.39}$ & \textbf{47.42}$_{\pm19.59}$ & \underline{52.52}$_{\pm17.71}$ & \textbf{37.19}$_{\pm33.37}$ & 0.00$_{\pm0.00}$ & \textbf{62.81}$_{\pm36.80}$ & 29.65$_{\pm23.44}$ \\
\cmidrule{2-11}
 & \multirow{5}{*}{JAMIE} & 500 & 26.87$_{\pm0.70}$ & 15.07$_{\pm0.71}$ & 25.59$_{\pm0.72}$ & 25.56$_{\pm0.95}$ & 5.58$_{\pm0.28}$ & 1.65$_{\pm0.77}$ & \textbf{8.29}$_{\pm8.61}$ & \textbf{52.29}$_{\pm2.83}$ \\
 &  & 1000 & 38.05$_{\pm0.83}$ & 23.43$_{\pm1.19}$ & 37.00$_{\pm0.85}$ & 35.82$_{\pm1.02}$ & 6.22$_{\pm0.14}$ & 1.03$_{\pm0.02}$ & 18.27$_{\pm1.07}$ & 54.80$_{\pm2.84}$ \\
 &  & 2000 & 47.68$_{\pm1.18}$ & 31.48$_{\pm1.54}$ & 46.79$_{\pm1.17}$ & 44.83$_{\pm1.46}$ & 7.24$_{\pm0.24}$ & \textbf{1.03}$_{\pm0.02}$ & 15.78$_{\pm0.48}$ & \underline{53.10}$_{\pm1.58}$ \\
 &  & 3000 & \underline{53.69}$_{\pm0.92}$ & \underline{37.58}$_{\pm1.92}$ & \underline{52.91}$_{\pm0.98}$ & \underline{49.79}$_{\pm0.85}$ & \underline{7.44}$_{\pm0.26}$ & \underline{1.03}$_{\pm0.03}$ & 15.48$_{\pm1.23}$ & 54.26$_{\pm2.62}$ \\
 &  & 4000 & \textbf{57.96}$_{\pm0.92}$ & \textbf{41.40}$_{\pm0.99}$ & \textbf{57.17}$_{\pm0.98}$ & \textbf{54.26}$_{\pm0.92}$ & \textbf{7.72}$_{\pm0.20}$ & 1.04$_{\pm0.04}$ & \underline{15.25}$_{\pm2.35}$ & 53.33$_{\pm2.25}$ \\
\cmidrule{2-11}
 & \multirow{5}{*}{scButterfly} & 500 & 61.73$_{\pm0.91}$ & 35.98$_{\pm3.43}$ & 60.28$_{\pm1.01}$ & 68.05$_{\pm1.03}$ & 30.07$_{\pm0.07}$ & \textbf{0.00}$_{\pm0.00}$ & 32.48$_{\pm2.84}$ & \textbf{48.52}$_{\pm8.61}$ \\
 &  & 1000 & 65.23$_{\pm0.57}$ & 38.74$_{\pm2.51}$ & 63.88$_{\pm0.64}$ & 72.26$_{\pm0.69}$ & 30.43$_{\pm0.11}$ & \underline{0.00}$_{\pm0.00}$ & 26.75$_{\pm2.83}$ & \underline{58.92}$_{\pm3.88}$ \\
 &  & 2000 & 66.47$_{\pm0.73}$ & 40.99$_{\pm1.82}$ & 65.09$_{\pm0.77}$ & 73.98$_{\pm1.26}$ & 30.67$_{\pm0.15}$ & 0.00$_{\pm0.00}$ & \underline{23.62}$_{\pm1.90}$ & 61.67$_{\pm3.32}$ \\
 &  & 3000 & \underline{66.87}$_{\pm1.36}$ & \underline{41.69}$_{\pm3.05}$ & \underline{65.56}$_{\pm1.45}$ & \underline{74.05}$_{\pm1.36}$ & \textbf{30.80}$_{\pm0.07}$ & 0.00$_{\pm0.00}$ & 23.72$_{\pm2.70}$ & 62.12$_{\pm3.89}$ \\
 &  & 4000 & \textbf{67.17}$_{\pm0.91}$ & \textbf{43.46}$_{\pm3.68}$ & \textbf{65.90}$_{\pm1.03}$ & \textbf{74.07}$_{\pm0.72}$ & \underline{30.77}$_{\pm0.08}$ & 0.00$_{\pm0.00}$ & \textbf{23.46}$_{\pm2.95}$ & 66.51$_{\pm1.70}$ \\
\cmidrule{2-11}
 & \multirow{5}{*}{multiDGD} & 500 & \textbf{15.65}$_{\pm15.74}$ & \textbf{8.06}$_{\pm9.44}$ & \textbf{14.33}$_{\pm16.20}$ & \textbf{13.11}$_{\pm12.32}$ & \underline{55.36}$_{\pm0.25}$ & \textbf{0.00}$_{\pm0.00}$ & \underline{90.48}$_{\pm6.97}$ & 39.58$_{\pm7.04}$ \\
 &  & 1000 & \underline{14.71}$_{\pm12.77}$ & \underline{5.74}$_{\pm5.36}$ & \underline{13.73}$_{\pm12.69}$ & \underline{12.42}$_{\pm10.87}$ & \textbf{55.43}$_{\pm0.28}$ & \underline{0.00}$_{\pm0.00}$ & \textbf{87.92}$_{\pm2.06}$ & 37.61$_{\pm6.77}$ \\
 &  & 2000 & 7.83$_{\pm8.64}$ & 2.18$_{\pm2.83}$ & 6.82$_{\pm8.60}$ & 6.49$_{\pm7.06}$ & 55.10$_{\pm0.90}$ & 0.00$_{\pm0.00}$ & 105.68$_{\pm38.80}$ & 32.69$_{\pm18.43}$ \\
 &  & 3000 & 3.64$_{\pm5.44}$ & 1.34$_{\pm3.14}$ & 2.39$_{\pm5.73}$ & 3.03$_{\pm4.29}$ & 51.98$_{\pm2.80}$ & 0.00$_{\pm0.00}$ & 250.48$_{\pm122.23}$ & \underline{8.96}$_{\pm20.03}$ \\
 &  & 4000 & 5.21$_{\pm5.41}$ & 1.69$_{\pm2.45}$ & 4.03$_{\pm5.73}$ & 4.33$_{\pm4.30}$ & 50.68$_{\pm2.12}$ & 0.00$_{\pm0.00}$ & 307.06$_{\pm83.21}$ & \textbf{0.00}$_{\pm0.00}$ \\
\cmidrule{2-11}
 & \multirow{5}{*}{scDiffusionX} & 500 & 49.98$_{\pm2.86}$ & 33.27$_{\pm2.89}$ & 49.24$_{\pm2.84}$ & 45.09$_{\pm3.44}$ & 29.64$_{\pm2.12}$ & 0.33$_{\pm0.01}$ & 318.51$_{\pm18.16}$ & \textbf{0.00}$_{\pm0.00}$ \\
 &  & 1000 & \underline{54.97}$_{\pm1.91}$ & \underline{37.78}$_{\pm2.54}$ & \underline{54.29}$_{\pm1.95}$ & \underline{49.56}$_{\pm1.83}$ & 29.80$_{\pm0.80}$ & 0.33$_{\pm0.01}$ & 319.35$_{\pm8.64}$ & \underline{0.00}$_{\pm0.00}$ \\
 &  & 2000 & \textbf{55.03}$_{\pm2.43}$ & \textbf{38.34}$_{\pm2.50}$ & \textbf{54.32}$_{\pm2.43}$ & \textbf{50.32}$_{\pm2.58}$ & \underline{30.26}$_{\pm1.56}$ & \underline{0.33}$_{\pm0.01}$ & \underline{315.27}$_{\pm13.73}$ & 0.00$_{\pm0.00}$ \\
 &  & 3000 & 53.42$_{\pm2.09}$ & 34.31$_{\pm3.05}$ & 52.71$_{\pm2.20}$ & 47.85$_{\pm1.33}$ & \textbf{31.53}$_{\pm1.34}$ & \textbf{0.33}$_{\pm0.01}$ & \textbf{303.10}$_{\pm14.67}$ & 0.00$_{\pm0.00}$ \\
 &  & 4000 & 53.34$_{\pm3.12}$ & 36.09$_{\pm3.71}$ & 52.67$_{\pm3.17}$ & 48.14$_{\pm2.88}$ & 29.83$_{\pm1.27}$ & 0.33$_{\pm0.01}$ & 318.91$_{\pm8.92}$ & 0.00$_{\pm0.00}$ \\
\midrule
\multirow{30}{*}{PBMC} & \multirow{5}{*}{BABEL} & 500 & 68.26$_{\pm1.30}$ & 41.45$_{\pm4.54}$ & 67.35$_{\pm1.38}$ & 71.75$_{\pm1.02}$ & 22.09$_{\pm0.07}$ & \textbf{0.02}$_{\pm0.00}$ & 7.81$_{\pm1.27}$ & \textbf{58.63}$_{\pm1.89}$ \\
 &  & 1000 & 71.29$_{\pm1.46}$ & \underline{43.95}$_{\pm3.94}$ & 70.35$_{\pm1.52}$ & 77.16$_{\pm0.76}$ & 22.31$_{\pm0.09}$ & \underline{0.02}$_{\pm0.00}$ & 5.47$_{\pm1.03}$ & \underline{67.46}$_{\pm3.50}$ \\
 &  & 2000 & 71.26$_{\pm1.52}$ & 41.92$_{\pm2.70}$ & 70.30$_{\pm1.58}$ & 77.94$_{\pm2.00}$ & 22.35$_{\pm0.05}$ & 0.02$_{\pm0.00}$ & \underline{3.89}$_{\pm0.58}$ & 75.20$_{\pm2.56}$ \\
 &  & 3000 & \underline{71.94}$_{\pm2.04}$ & 42.93$_{\pm5.76}$ & \underline{71.01}$_{\pm2.13}$ & \textbf{78.14}$_{\pm1.10}$ & \textbf{22.39}$_{\pm0.12}$ & 0.02$_{\pm0.00}$ & 5.77$_{\pm2.24}$ & 72.47$_{\pm2.26}$ \\
 &  & 4000 & \textbf{72.02}$_{\pm1.54}$ & \textbf{44.79}$_{\pm3.38}$ & \textbf{71.12}$_{\pm1.61}$ & \underline{78.11}$_{\pm1.74}$ & \underline{22.38}$_{\pm0.09}$ & 0.02$_{\pm0.00}$ & \textbf{3.51}$_{\pm0.99}$ & 75.51$_{\pm1.61}$ \\
\cmidrule{2-11}
 & \multirow{5}{*}{scPair} & 500 & 73.94$_{\pm0.90}$ & 51.28$_{\pm3.01}$ & 73.14$_{\pm0.93}$ & 77.65$_{\pm1.34}$ & 63.19$_{\pm0.42}$ & 29.07$_{\pm18.72}$ & \textbf{20.59}$_{\pm8.40}$ & \textbf{0.00}$_{\pm0.00}$ \\
 &  & 1000 & \underline{78.21}$_{\pm1.48}$ & \underline{59.16}$_{\pm4.36}$ & \underline{77.58}$_{\pm1.56}$ & \underline{81.19}$_{\pm1.39}$ & 68.01$_{\pm0.46}$ & 15.38$_{\pm9.33}$ & \underline{35.68}$_{\pm12.82}$ & \underline{0.00}$_{\pm0.00}$ \\
 &  & 2000 & \textbf{78.78}$_{\pm1.17}$ & \textbf{60.55}$_{\pm3.43}$ & \textbf{78.17}$_{\pm1.24}$ & \textbf{81.96}$_{\pm0.77}$ & \underline{69.76}$_{\pm0.22}$ & \underline{8.05}$_{\pm4.66}$ & 45.70$_{\pm15.82}$ & 0.00$_{\pm0.00}$ \\
 &  & 3000 & 68.26$_{\pm1.25}$ & 42.41$_{\pm4.73}$ & 67.21$_{\pm1.32}$ & 74.13$_{\pm0.91}$ & 53.10$_{\pm0.13}$ & \textbf{0.68}$_{\pm0.01}$ & 316.47$_{\pm1.46}$ & 0.00$_{\pm0.00}$ \\
 &  & 4000 & 64.91$_{\pm2.26}$ & 37.45$_{\pm2.77}$ & 63.86$_{\pm2.34}$ & 69.28$_{\pm2.49}$ & \textbf{96.64}$_{\pm0.10}$ & 15.90$_{\pm2.42}$ & 420.34$_{\pm25.98}$ & 0.00$_{\pm0.00}$ \\
\cmidrule{2-11}
 & \multirow{5}{*}{JAMIE} & 500 & 63.80$_{\pm1.02}$ & 42.99$_{\pm3.59}$ & 62.69$_{\pm1.05}$ & 67.94$_{\pm1.18}$ & 13.55$_{\pm0.10}$ & 5.82$_{\pm5.24}$ & \textbf{0.78}$_{\pm0.95}$ & 74.71$_{\pm1.56}$ \\
 &  & 1000 & 66.20$_{\pm1.08}$ & 43.26$_{\pm2.30}$ & 65.14$_{\pm1.15}$ & 70.99$_{\pm1.24}$ & 14.10$_{\pm0.15}$ & 0.99$_{\pm0.01}$ & \underline{5.40}$_{\pm0.67}$ & 71.88$_{\pm2.37}$ \\
 &  & 2000 & 69.81$_{\pm1.68}$ & 47.70$_{\pm3.19}$ & 68.89$_{\pm1.73}$ & 74.08$_{\pm1.68}$ & 14.41$_{\pm0.12}$ & 0.99$_{\pm0.01}$ & 5.60$_{\pm0.49}$ & 73.11$_{\pm2.32}$ \\
 &  & 3000 & \underline{71.80}$_{\pm1.05}$ & \textbf{53.51}$_{\pm5.40}$ & \underline{70.93}$_{\pm1.09}$ & \underline{75.68}$_{\pm0.66}$ & \underline{14.60}$_{\pm0.14}$ & \underline{0.98}$_{\pm0.01}$ & 6.28$_{\pm0.20}$ & \underline{70.02}$_{\pm1.45}$ \\
 &  & 4000 & \textbf{72.29}$_{\pm1.67}$ & \underline{49.76}$_{\pm5.73}$ & \textbf{71.42}$_{\pm1.68}$ & \textbf{77.18}$_{\pm1.45}$ & \textbf{14.73}$_{\pm0.10}$ & \textbf{0.98}$_{\pm0.01}$ & 6.82$_{\pm1.04}$ & \textbf{69.47}$_{\pm2.08}$ \\
\cmidrule{2-11}
 & \multirow{5}{*}{scButterfly} & 500 & 67.86$_{\pm1.11}$ & 33.04$_{\pm3.12}$ & 66.42$_{\pm1.29}$ & 79.15$_{\pm0.71}$ & 30.70$_{\pm0.08}$ & \textbf{0.00}$_{\pm0.00}$ & \textbf{2.75}$_{\pm0.37}$ & 39.70$_{\pm7.47}$ \\
 &  & 1000 & 71.39$_{\pm0.78}$ & 35.78$_{\pm1.53}$ & 70.10$_{\pm0.83}$ & 83.42$_{\pm0.48}$ & 30.78$_{\pm0.10}$ & \underline{0.00}$_{\pm0.00}$ & \underline{3.13}$_{\pm2.74}$ & 43.26$_{\pm16.14}$ \\
 &  & 2000 & \underline{74.25}$_{\pm0.99}$ & 39.79$_{\pm1.83}$ & \underline{73.12}$_{\pm1.08}$ & \underline{85.86}$_{\pm0.82}$ & \underline{30.84}$_{\pm0.08}$ & 0.00$_{\pm0.00}$ & 5.74$_{\pm1.22}$ & \underline{31.23}$_{\pm8.35}$ \\
 &  & 3000 & 74.15$_{\pm3.15}$ & \textbf{41.93}$_{\pm2.65}$ & 73.09$_{\pm3.17}$ & 84.64$_{\pm5.37}$ & 28.41$_{\pm5.47}$ & 0.00$_{\pm0.00}$ & 152.59$_{\pm329.95}$ & \textbf{27.91}$_{\pm15.92}$ \\
 &  & 4000 & \textbf{74.58}$_{\pm1.16}$ & \underline{40.03}$_{\pm2.68}$ & \textbf{73.50}$_{\pm1.25}$ & \textbf{85.97}$_{\pm0.75}$ & \textbf{30.87}$_{\pm0.09}$ & 0.00$_{\pm0.00}$ & 5.23$_{\pm0.74}$ & 33.09$_{\pm5.40}$ \\
\cmidrule{2-11}
 & \multirow{5}{*}{multiDGD} & 500 & 55.03$_{\pm2.40}$ & 20.72$_{\pm1.94}$ & 52.99$_{\pm2.54}$ & 64.74$_{\pm2.90}$ & \textbf{54.37}$_{\pm1.29}$ & \textbf{0.81}$_{\pm0.01}$ & \textbf{424.18}$_{\pm1.71}$ & \textbf{0.00}$_{\pm0.00}$ \\
 &  & 1000 & \textbf{55.55}$_{\pm1.24}$ & \underline{20.78}$_{\pm1.40}$ & \textbf{53.54}$_{\pm1.33}$ & \underline{65.18}$_{\pm1.20}$ & 53.84$_{\pm0.81}$ & \underline{0.81}$_{\pm0.01}$ & \underline{424.18}$_{\pm1.71}$ & \underline{0.00}$_{\pm0.00}$ \\
 &  & 2000 & \underline{55.46}$_{\pm2.63}$ & \textbf{20.96}$_{\pm1.53}$ & \underline{53.43}$_{\pm2.77}$ & \textbf{65.22}$_{\pm3.04}$ & 53.59$_{\pm0.88}$ & 0.81$_{\pm0.01}$ & 424.18$_{\pm1.71}$ & 0.00$_{\pm0.00}$ \\
 &  & 3000 & 53.49$_{\pm1.01}$ & 19.56$_{\pm1.56}$ & 51.33$_{\pm1.13}$ & 63.19$_{\pm0.64}$ & \underline{54.04}$_{\pm1.35}$ & 0.81$_{\pm0.01}$ & 424.18$_{\pm1.71}$ & 0.00$_{\pm0.00}$ \\
 &  & 4000 & 54.20$_{\pm1.18}$ & 20.14$_{\pm1.37}$ & 52.06$_{\pm1.33}$ & 64.19$_{\pm1.38}$ & 53.12$_{\pm0.97}$ & 0.81$_{\pm0.01}$ & 424.18$_{\pm1.71}$ & 0.00$_{\pm0.00}$ \\
\cmidrule{2-11}
 & \multirow{5}{*}{scDiffusionX} & 500 & 31.40$_{\pm2.06}$ & 12.56$_{\pm0.90}$ & 29.41$_{\pm2.09}$ & 32.03$_{\pm2.10}$ & 42.86$_{\pm0.12}$ & 0.12$_{\pm0.00}$ & \underline{192.41}$_{\pm6.55}$ & \textbf{0.00}$_{\pm0.00}$ \\
 &  & 1000 & 33.94$_{\pm2.90}$ & 16.12$_{\pm1.41}$ & 32.03$_{\pm2.96}$ & 35.45$_{\pm3.21}$ & 42.91$_{\pm0.30}$ & \underline{0.12}$_{\pm0.00}$ & 202.70$_{\pm9.75}$ & \underline{0.00}$_{\pm0.00}$ \\
 &  & 2000 & 35.38$_{\pm4.30}$ & 17.57$_{\pm2.74}$ & 33.49$_{\pm4.32}$ & 37.12$_{\pm4.99}$ & 43.02$_{\pm0.28}$ & 0.12$_{\pm0.00}$ & 198.43$_{\pm7.79}$ & 0.00$_{\pm0.00}$ \\
 &  & 3000 & \textbf{37.64}$_{\pm2.27}$ & \textbf{18.51}$_{\pm1.95}$ & \textbf{35.80}$_{\pm2.35}$ & \textbf{39.44}$_{\pm2.75}$ & \textbf{43.09}$_{\pm0.26}$ & \textbf{0.12}$_{\pm0.00}$ & \textbf{190.85}$_{\pm5.69}$ & 0.00$_{\pm0.00}$ \\
 &  & 4000 & \underline{37.00}$_{\pm2.59}$ & \underline{17.84}$_{\pm0.67}$ & \underline{35.12}$_{\pm2.66}$ & \underline{38.92}$_{\pm2.78}$ & \underline{43.04}$_{\pm0.15}$ & 0.12$_{\pm0.00}$ & 198.13$_{\pm5.30}$ & 0.00$_{\pm0.00}$ \\
\bottomrule
\end{tabular}
\end{table*}

\clearpage

\begin{table*}[htbp]
\centering
\caption{The table below presents the translation results from RNA to ATAC on \underline{\textbf{feature quality}} factor.}
\label{tab:sparse_results_r2a}
\scriptsize
\renewcommand{\arraystretch}{0.3}
\setlength{\tabcolsep}{3pt}
\begin{tabular}{lll|cccc|cc|cc}
\toprule
\multirow{2}{*}{\textbf{Dataset}} & \multirow{2}{*}{\textbf{Model}} & \multirow{2}{*}{\textbf{Sparse Ratio}} & \multicolumn{4}{c|}{\textbf{Clustering}} & \multicolumn{2}{c|}{\textbf{Regression}} & \multicolumn{2}{c}{\textbf{Distribution}} \\
\cmidrule(lr){4-7} \cmidrule(lr){8-9} \cmidrule(lr){10-11}
 & & & \textbf{NMI} & \textbf{ARI} & \textbf{AMI} & \textbf{HOM} & \textbf{PCC} & \textbf{MSE} & \textbf{MMD} & \textbf{LISI} \\
\midrule
\multirow{30}{*}{Brain} & \multirow{5}{*}{BABEL} & 0\% & \textbf{47.21}$_{\pm1.53}$ & \textbf{21.68}$_{\pm1.65}$ & \textbf{45.40}$_{\pm1.61}$ & \textbf{51.40}$_{\pm1.62}$ & \textbf{52.55}$_{\pm0.23}$ & 0.19$_{\pm0.00}$ & \underline{45.12}$_{\pm9.84}$ & 63.46$_{\pm3.65}$ \\
 &  & 20\% & \underline{45.30}$_{\pm1.96}$ & 20.25$_{\pm2.03}$ & \underline{43.05}$_{\pm2.05}$ & \underline{48.89}$_{\pm2.02}$ & \underline{40.59}$_{\pm0.19}$ & 0.21$_{\pm0.00}$ & \textbf{42.98}$_{\pm2.49}$ & 65.64$_{\pm1.63}$ \\
 &  & 40\% & 43.76$_{\pm1.98}$ & 19.15$_{\pm1.93}$ & 41.59$_{\pm2.03}$ & 46.52$_{\pm2.37}$ & 31.62$_{\pm0.07}$ & 0.20$_{\pm0.00}$ & 65.16$_{\pm7.63}$ & 61.37$_{\pm2.60}$ \\
 &  & 60\% & 43.34$_{\pm2.31}$ & \underline{20.33}$_{\pm1.92}$ & 41.35$_{\pm2.19}$ & 45.16$_{\pm3.11}$ & 23.48$_{\pm0.09}$ & \underline{0.16}$_{\pm0.00}$ & 96.79$_{\pm6.28}$ & \underline{57.20}$_{\pm1.78}$ \\
 &  & 80\% & 33.28$_{\pm2.02}$ & 16.31$_{\pm1.01}$ & 31.40$_{\pm1.90}$ & 33.03$_{\pm2.28}$ & 14.72$_{\pm0.06}$ & \textbf{0.10}$_{\pm0.00}$ & 113.52$_{\pm1.85}$ & \textbf{46.30}$_{\pm2.33}$ \\
\cmidrule{2-11}
 & \multirow{5}{*}{scPair} & 0\% & \textbf{48.63}$_{\pm21.27}$ & \textbf{27.31}$_{\pm23.33}$ & \textbf{46.35}$_{\pm22.60}$ & \textbf{52.78}$_{\pm19.71}$ & \textbf{31.62}$_{\pm23.93}$ & \textbf{0.00}$_{\pm0.00}$ & 80.03$_{\pm29.58}$ & 22.66$_{\pm30.70}$ \\
 &  & 20\% & \underline{21.18}$_{\pm2.25}$ & \underline{5.36}$_{\pm0.52}$ & \underline{17.20}$_{\pm2.05}$ & \underline{24.75}$_{\pm3.08}$ & 1.55$_{\pm7.12}$ & \underline{0.00}$_{\pm0.00}$ & \underline{79.42}$_{\pm87.27}$ & \underline{0.78}$_{\pm1.30}$ \\
 &  & 40\% & 14.80$_{\pm6.84}$ & 3.17$_{\pm2.40}$ & 10.43$_{\pm7.33}$ & 17.17$_{\pm7.79}$ & -2.40$_{\pm5.75}$ & 0.00$_{\pm0.00}$ & 195.30$_{\pm66.17}$ & \textbf{0.00}$_{\pm0.00}$ \\
 &  & 60\% & 14.48$_{\pm2.61}$ & 3.18$_{\pm0.94}$ & 10.53$_{\pm2.50}$ & 16.45$_{\pm3.15}$ & 4.27$_{\pm2.99}$ & 0.00$_{\pm0.00}$ & 146.97$_{\pm34.89}$ & 5.01$_{\pm6.62}$ \\
 &  & 80\% & 10.40$_{\pm3.59}$ & 1.83$_{\pm1.26}$ & 6.35$_{\pm3.40}$ & 11.88$_{\pm4.26}$ & \underline{9.18}$_{\pm2.51}$ & 0.00$_{\pm0.00}$ & \textbf{1.76}$_{\pm3.30}$ & 4.83$_{\pm6.94}$ \\
\cmidrule{2-11}
 & \multirow{5}{*}{JAMIE} & 0\% & \textbf{53.69}$_{\pm0.92}$ & \textbf{37.58}$_{\pm1.92}$ & \textbf{52.91}$_{\pm0.98}$ & \textbf{49.79}$_{\pm0.85}$ & \textbf{7.44}$_{\pm0.26}$ & 1.03$_{\pm0.03}$ & 15.48$_{\pm1.23}$ & 54.26$_{\pm2.62}$ \\
 &  & 20\% & \underline{51.75}$_{\pm0.96}$ & \underline{34.71}$_{\pm0.34}$ & \underline{50.90}$_{\pm0.94}$ & \underline{48.87}$_{\pm1.37}$ & \underline{5.96}$_{\pm0.24}$ & 1.01$_{\pm0.02}$ & 13.17$_{\pm1.07}$ & 51.14$_{\pm3.51}$ \\
 &  & 40\% & 47.43$_{\pm1.60}$ & 31.85$_{\pm1.74}$ & 46.43$_{\pm1.64}$ & 45.29$_{\pm1.38}$ & 4.55$_{\pm0.18}$ & 1.01$_{\pm0.02}$ & \textbf{12.28}$_{\pm0.73}$ & 50.33$_{\pm2.51}$ \\
 &  & 60\% & 36.43$_{\pm1.80}$ & 21.24$_{\pm1.21}$ & 35.38$_{\pm1.84}$ & 34.34$_{\pm1.47}$ & 3.01$_{\pm0.13}$ & \underline{1.00}$_{\pm0.02}$ & 13.64$_{\pm1.52}$ & \underline{44.69}$_{\pm2.06}$ \\
 &  & 80\% & 26.63$_{\pm1.09}$ & 14.67$_{\pm0.45}$ & 25.34$_{\pm1.15}$ & 25.34$_{\pm1.10}$ & 1.39$_{\pm0.10}$ & \textbf{1.00}$_{\pm0.01}$ & \underline{12.94}$_{\pm1.22}$ & \textbf{28.35}$_{\pm3.60}$ \\
\cmidrule{2-11}
 & \multirow{5}{*}{scButterfly} & 0\% & \underline{66.87}$_{\pm1.36}$ & \underline{41.69}$_{\pm3.05}$ & \underline{65.56}$_{\pm1.45}$ & \underline{74.05}$_{\pm1.36}$ & \textbf{30.80}$_{\pm0.07}$ & \textbf{0.00}$_{\pm0.00}$ & 23.72$_{\pm2.70}$ & 62.12$_{\pm3.89}$ \\
 &  & 20\% & \textbf{67.60}$_{\pm0.90}$ & 40.58$_{\pm1.52}$ & \textbf{66.27}$_{\pm0.95}$ & \textbf{75.62}$_{\pm0.99}$ & \underline{25.27}$_{\pm0.06}$ & \underline{0.00}$_{\pm0.00}$ & \textbf{20.09}$_{\pm1.89}$ & 64.23$_{\pm1.79}$ \\
 &  & 40\% & 65.88$_{\pm1.62}$ & \textbf{41.85}$_{\pm3.12}$ & 64.65$_{\pm1.71}$ & 72.04$_{\pm1.69}$ & 20.60$_{\pm0.06}$ & 0.00$_{\pm0.00}$ & 27.00$_{\pm2.01}$ & 51.81$_{\pm5.94}$ \\
 &  & 60\% & 60.10$_{\pm1.85}$ & 39.21$_{\pm3.69}$ & 58.87$_{\pm1.92}$ & 63.88$_{\pm1.86}$ & 16.10$_{\pm0.06}$ & 0.00$_{\pm0.00}$ & 26.73$_{\pm1.57}$ & \underline{48.16}$_{\pm3.11}$ \\
 &  & 80\% & 32.99$_{\pm7.14}$ & 13.54$_{\pm4.38}$ & 30.58$_{\pm7.59}$ & 36.05$_{\pm7.10}$ & 10.95$_{\pm0.04}$ & 0.00$_{\pm0.00}$ & \underline{23.28}$_{\pm0.52}$ & \textbf{27.32}$_{\pm11.03}$ \\
\cmidrule{2-11}
 & \multirow{5}{*}{multiDGD} & 0\% & \textbf{3.64}$_{\pm5.44}$ & \textbf{1.34}$_{\pm3.14}$ & \textbf{2.39}$_{\pm5.73}$ & \textbf{3.03}$_{\pm4.29}$ & \textbf{51.98}$_{\pm2.80}$ & \textbf{0.00}$_{\pm0.00}$ & \underline{250.48}$_{\pm122.23}$ & 8.96$_{\pm20.03}$ \\
 &  & 20\% & 1.47$_{\pm0.70}$ & 0.09$_{\pm0.06}$ & 0.06$_{\pm0.21}$ & 1.36$_{\pm0.74}$ & \underline{46.85}$_{\pm0.98}$ & \underline{0.00}$_{\pm0.00}$ & \textbf{248.78}$_{\pm2.45}$ & \textbf{0.00}$_{\pm0.00}$ \\
 &  & 40\% & 2.29$_{\pm0.69}$ & 0.00$_{\pm0.16}$ & 0.32$_{\pm0.54}$ & 2.20$_{\pm0.72}$ & 38.18$_{\pm1.92}$ & 0.00$_{\pm0.00}$ & 456.14$_{\pm10.45}$ & \underline{0.00}$_{\pm0.00}$ \\
 &  & 60\% & 2.32$_{\pm1.07}$ & \underline{0.20}$_{\pm0.23}$ & \underline{0.60}$_{\pm0.80}$ & 2.21$_{\pm1.11}$ & 30.34$_{\pm0.36}$ & 0.00$_{\pm0.00}$ & 601.54$_{\pm11.52}$ & 0.00$_{\pm0.00}$ \\
 &  & 80\% & \underline{2.36}$_{\pm0.94}$ & 0.01$_{\pm0.27}$ & 0.41$_{\pm1.19}$ & \underline{2.26}$_{\pm0.81}$ & 20.35$_{\pm0.17}$ & 0.00$_{\pm0.00}$ & 689.90$_{\pm5.54}$ & 0.00$_{\pm0.00}$ \\
\cmidrule{2-11}
 & \multirow{5}{*}{scDiffusionX} & 0\% & \textbf{53.42}$_{\pm2.09}$ & \textbf{34.31}$_{\pm3.05}$ & \textbf{52.71}$_{\pm2.20}$ & \textbf{47.85}$_{\pm1.33}$ & \textbf{31.53}$_{\pm1.34}$ & \textbf{0.33}$_{\pm0.01}$ & 303.10$_{\pm14.67}$ & \textbf{0.00}$_{\pm0.00}$ \\
 &  & 20\% & \underline{46.92}$_{\pm1.17}$ & \underline{31.42}$_{\pm1.49}$ & \underline{45.98}$_{\pm1.17}$ & \underline{44.09}$_{\pm1.37}$ & \underline{28.92}$_{\pm0.03}$ & \underline{0.50}$_{\pm0.01}$ & \underline{25.68}$_{\pm1.43}$ & 48.69$_{\pm2.08}$ \\
 &  & 40\% & 43.26$_{\pm2.89}$ & 26.72$_{\pm2.56}$ & 42.42$_{\pm3.02}$ & 38.99$_{\pm1.93}$ & 19.27$_{\pm0.39}$ & 0.81$_{\pm0.03}$ & 66.78$_{\pm96.02}$ & 38.82$_{\pm21.77}$ \\
 &  & 60\% & 34.58$_{\pm3.64}$ & 21.22$_{\pm2.70}$ & 33.81$_{\pm3.71}$ & 29.37$_{\pm3.11}$ & 11.31$_{\pm0.10}$ & 1.46$_{\pm0.03}$ & 72.17$_{\pm112.58}$ & \underline{36.85}$_{\pm20.70}$ \\
 &  & 80\% & 20.33$_{\pm1.07}$ & 13.65$_{\pm0.74}$ & 19.15$_{\pm1.15}$ & 18.54$_{\pm1.04}$ & 4.90$_{\pm0.03}$ & 3.41$_{\pm0.04}$ & \textbf{19.36}$_{\pm0.89}$ & 44.65$_{\pm5.32}$ \\
\midrule
\multirow{30}{*}{PBMC} & \multirow{5}{*}{BABEL} & 0\% & \textbf{71.94}$_{\pm2.04}$ & \textbf{42.93}$_{\pm5.76}$ & \textbf{71.01}$_{\pm2.13}$ & \textbf{78.14}$_{\pm1.10}$ & \textbf{22.39}$_{\pm0.12}$ & 0.02$_{\pm0.00}$ & \textbf{5.77}$_{\pm2.24}$ & 72.47$_{\pm2.26}$ \\
 &  & 20\% & \underline{67.13}$_{\pm6.20}$ & \underline{40.36}$_{\pm8.04}$ & \underline{65.94}$_{\pm6.24}$ & \underline{72.35}$_{\pm6.62}$ & \underline{18.86}$_{\pm1.01}$ & 0.02$_{\pm0.00}$ & \underline{9.44}$_{\pm9.94}$ & 57.22$_{\pm32.05}$ \\
 &  & 40\% & 63.91$_{\pm7.43}$ & 35.96$_{\pm8.69}$ & 62.58$_{\pm7.64}$ & 69.00$_{\pm6.87}$ & 15.93$_{\pm0.80}$ & 0.02$_{\pm0.00}$ & 9.66$_{\pm8.38}$ & 55.67$_{\pm31.15}$ \\
 &  & 60\% & 59.52$_{\pm9.00}$ & 33.10$_{\pm7.67}$ & 58.14$_{\pm9.21}$ & 63.31$_{\pm8.79}$ & 12.26$_{\pm1.03}$ & \underline{0.01}$_{\pm0.00}$ & 11.75$_{\pm9.17}$ & \underline{52.04}$_{\pm28.25}$ \\
 &  & 80\% & 51.28$_{\pm7.58}$ & 29.47$_{\pm7.69}$ & 49.96$_{\pm7.75}$ & 51.84$_{\pm6.87}$ & 7.91$_{\pm0.75}$ & \textbf{0.01}$_{\pm0.00}$ & 17.82$_{\pm9.53}$ & \textbf{44.27}$_{\pm24.18}$ \\
\cmidrule{2-11}
 & \multirow{5}{*}{scPair} & 0\% & \underline{68.26}$_{\pm1.25}$ & \underline{42.41}$_{\pm4.73}$ & \underline{67.21}$_{\pm1.32}$ & \textbf{74.13}$_{\pm0.91}$ & \textbf{53.10}$_{\pm0.13}$ & 0.68$_{\pm0.01}$ & 316.47$_{\pm1.46}$ & \textbf{0.00}$_{\pm0.00}$ \\
 &  & 20\% & \textbf{69.01}$_{\pm1.80}$ & \textbf{44.13}$_{\pm2.98}$ & \textbf{68.05}$_{\pm1.82}$ & \underline{74.02}$_{\pm2.12}$ & \underline{46.99}$_{\pm0.18}$ & 0.58$_{\pm0.00}$ & 298.10$_{\pm0.65}$ & \underline{0.00}$_{\pm0.00}$ \\
 &  & 40\% & 59.50$_{\pm1.50}$ & 31.43$_{\pm2.28}$ & 58.30$_{\pm1.56}$ & 63.57$_{\pm1.10}$ & 40.24$_{\pm0.11}$ & 0.50$_{\pm0.00}$ & 270.76$_{\pm1.67}$ & 0.00$_{\pm0.00}$ \\
 &  & 60\% & 56.88$_{\pm1.77}$ & 32.43$_{\pm2.31}$ & 55.69$_{\pm1.88}$ & 59.56$_{\pm0.83}$ & 32.28$_{\pm0.08}$ & \underline{0.42}$_{\pm0.00}$ & \underline{234.14}$_{\pm1.80}$ & 0.00$_{\pm0.00}$ \\
 &  & 80\% & 43.81$_{\pm1.09}$ & 19.40$_{\pm1.21}$ & 42.15$_{\pm1.16}$ & 46.94$_{\pm1.13}$ & 21.48$_{\pm0.03}$ & \textbf{0.32}$_{\pm0.00}$ & \textbf{183.68}$_{\pm1.78}$ & 0.21$_{\pm0.21}$ \\
\cmidrule{2-11}
 & \multirow{5}{*}{JAMIE} & 0\% & \textbf{71.80}$_{\pm1.05}$ & 53.51$_{\pm5.40}$ & \textbf{70.93}$_{\pm1.09}$ & \textbf{75.68}$_{\pm0.66}$ & \textbf{14.60}$_{\pm0.14}$ & \textbf{0.98}$_{\pm0.01}$ & 6.28$_{\pm0.20}$ & 70.02$_{\pm1.45}$ \\
 &  & 20\% & \underline{69.50}$_{\pm0.55}$ & 51.03$_{\pm4.05}$ & \underline{68.57}$_{\pm0.57}$ & \underline{73.62}$_{\pm1.57}$ & \underline{12.45}$_{\pm0.10}$ & \underline{0.99}$_{\pm0.01}$ & \textbf{6.10}$_{\pm0.77}$ & 66.27$_{\pm3.49}$ \\
 &  & 40\% & 68.03$_{\pm1.94}$ & \underline{57.03}$_{\pm9.39}$ & 67.08$_{\pm1.98}$ & 70.77$_{\pm1.67}$ & 10.67$_{\pm0.06}$ & 0.99$_{\pm0.01}$ & \underline{6.18}$_{\pm0.66}$ & 64.08$_{\pm1.49}$ \\
 &  & 60\% & 64.06$_{\pm1.78}$ & \textbf{60.62}$_{\pm9.46}$ & 62.99$_{\pm1.83}$ & 65.79$_{\pm1.23}$ & 8.66$_{\pm0.12}$ & 1.00$_{\pm0.01}$ & 6.93$_{\pm0.81}$ & \underline{62.26}$_{\pm2.06}$ \\
 &  & 80\% & 56.90$_{\pm1.54}$ & 53.38$_{\pm3.80}$ & 55.86$_{\pm1.58}$ & 56.11$_{\pm0.80}$ & 5.42$_{\pm0.06}$ & 1.01$_{\pm0.01}$ & 8.63$_{\pm0.88}$ & \textbf{55.64}$_{\pm2.66}$ \\
\cmidrule{2-11}
 & \multirow{5}{*}{scButterfly} & 0\% & \textbf{74.15}$_{\pm3.15}$ & \textbf{41.93}$_{\pm2.65}$ & \textbf{73.09}$_{\pm3.17}$ & \textbf{84.64}$_{\pm5.37}$ & \textbf{28.41}$_{\pm5.47}$ & 0.00$_{\pm0.00}$ & 152.59$_{\pm329.95}$ & \underline{27.91}$_{\pm15.92}$ \\
 &  & 20\% & \underline{71.08}$_{\pm1.62}$ & \underline{35.49}$_{\pm2.78}$ & \underline{69.80}$_{\pm1.75}$ & \underline{82.85}$_{\pm1.14}$ & \underline{28.22}$_{\pm0.06}$ & \textbf{0.00}$_{\pm0.00}$ & \textbf{4.61}$_{\pm0.86}$ & 38.98$_{\pm4.55}$ \\
 &  & 40\% & 65.25$_{\pm1.73}$ & 30.82$_{\pm2.92}$ & 63.76$_{\pm1.89}$ & 75.62$_{\pm1.03}$ & 25.11$_{\pm0.06}$ & \underline{0.00}$_{\pm0.00}$ & \underline{4.66}$_{\pm0.80}$ & 37.30$_{\pm3.93}$ \\
 &  & 60\% & 57.20$_{\pm0.52}$ & 26.14$_{\pm1.49}$ & 55.60$_{\pm0.57}$ & 64.64$_{\pm0.79}$ & 21.05$_{\pm0.02}$ & 0.00$_{\pm0.00}$ & 6.58$_{\pm0.87}$ & 34.55$_{\pm2.94}$ \\
 &  & 80\% & 42.84$_{\pm3.71}$ & 16.85$_{\pm1.58}$ & 40.72$_{\pm3.76}$ & 48.51$_{\pm4.63}$ & 14.96$_{\pm0.08}$ & 0.00$_{\pm0.00}$ & 11.98$_{\pm2.09}$ & \textbf{23.34}$_{\pm13.27}$ \\
\cmidrule{2-11}
 & \multirow{5}{*}{multiDGD} & 0\% & \textbf{53.49}$_{\pm1.01}$ & \underline{19.56}$_{\pm1.56}$ & \textbf{51.33}$_{\pm1.13}$ & \underline{63.19}$_{\pm0.64}$ & \textbf{54.04}$_{\pm1.35}$ & 0.81$_{\pm0.01}$ & 424.18$_{\pm1.71}$ & \textbf{0.00}$_{\pm0.00}$ \\
 &  & 20\% & \underline{53.38}$_{\pm1.37}$ & \textbf{19.61}$_{\pm0.96}$ & \underline{51.09}$_{\pm1.38}$ & \textbf{63.79}$_{\pm1.83}$ & \underline{46.87}$_{\pm0.81}$ & 0.67$_{\pm0.01}$ & 400.34$_{\pm1.47}$ & \underline{0.00}$_{\pm0.00}$ \\
 &  & 40\% & 52.15$_{\pm1.56}$ & 18.75$_{\pm1.61}$ & 49.89$_{\pm1.62}$ & 61.90$_{\pm1.97}$ & 40.44$_{\pm0.26}$ & 0.56$_{\pm0.00}$ & 367.88$_{\pm1.64}$ & 0.00$_{\pm0.00}$ \\
 &  & 60\% & 51.88$_{\pm1.71}$ & 19.34$_{\pm1.73}$ & 49.74$_{\pm1.88}$ & 60.70$_{\pm1.74}$ & 33.23$_{\pm0.54}$ & \underline{0.45}$_{\pm0.00}$ & \underline{321.77}$_{\pm1.26}$ & 0.00$_{\pm0.00}$ \\
 &  & 80\% & 49.27$_{\pm1.93}$ & 17.37$_{\pm1.32}$ & 47.15$_{\pm1.99}$ & 57.03$_{\pm2.26}$ & 22.79$_{\pm0.16}$ & \textbf{0.33}$_{\pm0.00}$ & \textbf{250.38}$_{\pm1.44}$ & 0.00$_{\pm0.00}$ \\
\cmidrule{2-11}
 & \multirow{5}{*}{scDiffusionX} & 0\% & \textbf{37.64}$_{\pm2.27}$ & \underline{18.51}$_{\pm1.95}$ & \textbf{35.80}$_{\pm2.35}$ & \textbf{39.44}$_{\pm2.75}$ & \textbf{43.09}$_{\pm0.26}$ & \textbf{0.12}$_{\pm0.00}$ & 190.85$_{\pm5.69}$ & \textbf{0.00}$_{\pm0.00}$ \\
 &  & 20\% & \underline{31.97}$_{\pm2.61}$ & 16.13$_{\pm1.51}$ & \underline{30.25}$_{\pm2.75}$ & \underline{32.10}$_{\pm2.14}$ & \underline{37.03}$_{\pm0.21}$ & \underline{0.16}$_{\pm0.00}$ & \underline{179.20}$_{\pm4.66}$ & \underline{0.00}$_{\pm0.00}$ \\
 &  & 40\% & 26.40$_{\pm5.02}$ & 14.89$_{\pm4.23}$ & 25.15$_{\pm5.13}$ & 24.63$_{\pm4.77}$ & 29.38$_{\pm0.58}$ & 0.27$_{\pm0.00}$ & \textbf{178.13}$_{\pm11.64}$ & 0.00$_{\pm0.00}$ \\
 &  & 60\% & 21.18$_{\pm3.20}$ & 12.96$_{\pm1.93}$ & 20.03$_{\pm3.27}$ & 19.24$_{\pm2.93}$ & 20.38$_{\pm0.40}$ & 0.54$_{\pm0.00}$ & 194.65$_{\pm6.11}$ & 0.00$_{\pm0.00}$ \\
 &  & 80\% & 23.94$_{\pm2.64}$ & \textbf{21.19}$_{\pm2.81}$ & 23.03$_{\pm2.66}$ & 20.73$_{\pm2.23}$ & 9.82$_{\pm0.13}$ & 1.84$_{\pm0.02}$ & 186.25$_{\pm3.06}$ & 0.00$_{\pm0.00}$ \\
\bottomrule
\end{tabular}
\end{table*}

\clearpage

\begin{table*}[htbp]
\centering
\caption{The table below presents the translation results from RNA to ATAC on \underline{\textbf{few-shot learning}} factor.}
\label{tab:fewshot_results_r2a}
\scriptsize
\renewcommand{\arraystretch}{1.0}
\setlength{\tabcolsep}{4pt}
\begin{tabular}{lll|cccc|cc|cc}
\toprule
\multirow{2}{*}{\textbf{Dataset}} & \multirow{2}{*}{\textbf{Model}} & \multirow{2}{*}{\textbf{Kind}} & \multicolumn{4}{c|}{\textbf{Clustering}} & \multicolumn{2}{c|}{\textbf{Regression}} & \multicolumn{2}{c}{\textbf{Distribution}} \\
\cmidrule(lr){4-7} \cmidrule(lr){8-9} \cmidrule(lr){10-11}
 & & & \textbf{NMI} & \textbf{ARI} & \textbf{AMI} & \textbf{HOM} & \textbf{PCC} & \textbf{MSE} & \textbf{MMD} & \textbf{LISI} \\
\midrule
\multirow{12}{*}{Brain} & \multirow{2}{*}{BABEL} & fewshot & \underline{46.90}$_{\pm2.76}$ & \underline{20.95}$_{\pm3.82}$ & \textbf{45.41}$_{\pm2.09}$ & \textbf{51.78}$_{\pm1.73}$ & \underline{52.45}$_{\pm0.36}$ & \underline{0.19}$_{\pm0.00}$ & \textbf{34.19}$_{\pm16.64}$ & \textbf{58.79}$_{\pm5.16}$ \\
 &  & normal & \textbf{47.21}$_{\pm1.53}$ & \textbf{21.68}$_{\pm1.65}$ & \underline{45.40}$_{\pm1.61}$ & \underline{51.40}$_{\pm1.62}$ & \textbf{52.55}$_{\pm0.23}$ & \textbf{0.19}$_{\pm0.00}$ & \underline{45.12}$_{\pm9.84}$ & \underline{63.46}$_{\pm3.65}$ \\
\cmidrule{2-11}
 & \multirow{2}{*}{scPair} & fewshot & \underline{30.50}$_{\pm3.89}$ & \underline{8.74}$_{\pm1.30}$ & \underline{29.05}$_{\pm4.05}$ & \underline{38.97}$_{\pm4.71}$ & \underline{1.05}$_{\pm9.22}$ & \textbf{0.00}$_{\pm0.00}$ & \underline{83.38}$_{\pm60.06}$ & \textbf{0.00}$_{\pm0.00}$ \\
 &  & normal & \textbf{48.63}$_{\pm21.27}$ & \textbf{27.31}$_{\pm23.33}$ & \textbf{46.35}$_{\pm22.60}$ & \textbf{52.78}$_{\pm19.71}$ & \textbf{31.62}$_{\pm23.93}$ & \underline{0.00}$_{\pm0.00}$ & \textbf{80.03}$_{\pm29.58}$ & \underline{22.66}$_{\pm30.70}$ \\
\cmidrule{2-11}
 & \multirow{2}{*}{JAMIE} & fewshot & \underline{47.84}$_{\pm1.12}$ & \underline{33.47}$_{\pm1.13}$ & \underline{47.62}$_{\pm1.12}$ & \underline{44.45}$_{\pm1.02}$ & \underline{4.69}$_{\pm0.18}$ & \underline{1.05}$_{\pm0.01}$ & \underline{29.20}$_{\pm4.31}$ & \textbf{48.40}$_{\pm2.25}$ \\
 &  & normal & \textbf{53.69}$_{\pm0.92}$ & \textbf{37.58}$_{\pm1.92}$ & \textbf{52.91}$_{\pm0.98}$ & \textbf{49.79}$_{\pm0.85}$ & \textbf{7.44}$_{\pm0.26}$ & \textbf{1.03}$_{\pm0.03}$ & \textbf{15.48}$_{\pm1.23}$ & \underline{54.26}$_{\pm2.62}$ \\
\cmidrule{2-11}
 & \multirow{2}{*}{scButterfly} & fewshot & \underline{62.99}$_{\pm0.68}$ & \underline{33.71}$_{\pm1.54}$ & \underline{62.54}$_{\pm0.71}$ & \underline{73.67}$_{\pm0.42}$ & \underline{29.18}$_{\pm0.18}$ & \textbf{0.00}$_{\pm0.00}$ & \underline{46.89}$_{\pm5.35}$ & \textbf{28.54}$_{\pm4.31}$ \\
 &  & normal & \textbf{66.87}$_{\pm1.36}$ & \textbf{41.69}$_{\pm3.05}$ & \textbf{65.56}$_{\pm1.45}$ & \textbf{74.05}$_{\pm1.36}$ & \textbf{30.80}$_{\pm0.07}$ & \underline{0.00}$_{\pm0.00}$ & \textbf{23.72}$_{\pm2.70}$ & \underline{62.12}$_{\pm3.89}$ \\
\cmidrule{2-11}
 & \multirow{2}{*}{multiDGD} & fewshot & \underline{1.14}$_{\pm0.98}$ & \underline{0.07}$_{\pm0.15}$ & \underline{0.68}$_{\pm0.91}$ & \underline{1.05}$_{\pm0.93}$ & \textbf{52.61}$_{\pm0.07}$ & \textbf{0.00}$_{\pm0.00}$ & \underline{285.04}$_{\pm0.55}$ & \textbf{0.00}$_{\pm0.00}$ \\
 &  & normal & \textbf{3.64}$_{\pm5.44}$ & \textbf{1.34}$_{\pm3.14}$ & \textbf{2.39}$_{\pm5.73}$ & \textbf{3.03}$_{\pm4.29}$ & \underline{51.98}$_{\pm2.80}$ & \underline{0.00}$_{\pm0.00}$ & \textbf{250.48}$_{\pm122.23}$ & \underline{8.96}$_{\pm20.03}$ \\
\cmidrule{2-11}
 & \multirow{2}{*}{scDiffusionX} & fewshot & \underline{52.38}$_{\pm2.22}$ & \underline{34.25}$_{\pm2.15}$ & \underline{51.71}$_{\pm2.22}$ & \underline{46.44}$_{\pm2.20}$ & \textbf{32.59}$_{\pm1.27}$ & \textbf{0.32}$_{\pm0.01}$ & \textbf{294.46}$_{\pm11.99}$ & \textbf{0.00}$_{\pm0.00}$ \\
 &  & normal & \textbf{53.42}$_{\pm2.09}$ & \textbf{34.31}$_{\pm3.05}$ & \textbf{52.71}$_{\pm2.20}$ & \textbf{47.85}$_{\pm1.33}$ & \underline{31.53}$_{\pm1.34}$ & \underline{0.33}$_{\pm0.01}$ & \underline{303.10}$_{\pm14.67}$ & \underline{0.00}$_{\pm0.00}$ \\
\midrule
\multirow{12}{*}{PBMC} & \multirow{2}{*}{BABEL} & fewshot & \underline{67.84}$_{\pm0.80}$ & \underline{36.35}$_{\pm1.76}$ & \underline{67.52}$_{\pm0.82}$ & \underline{77.62}$_{\pm0.67}$ & \underline{21.57}$_{\pm0.07}$ & \underline{0.02}$_{\pm0.00}$ & \underline{11.79}$_{\pm0.42}$ & \textbf{0.22}$_{\pm0.21}$ \\
 &  & normal & \textbf{71.94}$_{\pm2.04}$ & \textbf{42.93}$_{\pm5.76}$ & \textbf{71.01}$_{\pm2.13}$ & \textbf{78.14}$_{\pm1.10}$ & \textbf{22.39}$_{\pm0.12}$ & \textbf{0.02}$_{\pm0.00}$ & \textbf{5.77}$_{\pm2.24}$ & \underline{72.47}$_{\pm2.26}$ \\
\cmidrule{2-11}
 & \multirow{2}{*}{scPair} & fewshot & \underline{56.61}$_{\pm2.82}$ & \underline{26.26}$_{\pm3.48}$ & \underline{56.16}$_{\pm2.87}$ & \underline{65.23}$_{\pm2.43}$ & \underline{51.50}$_{\pm0.28}$ & \underline{0.68}$_{\pm0.00}$ & \underline{329.57}$_{\pm1.60}$ & \textbf{0.00}$_{\pm0.00}$ \\
 &  & normal & \textbf{68.26}$_{\pm1.25}$ & \textbf{42.41}$_{\pm4.73}$ & \textbf{67.21}$_{\pm1.32}$ & \textbf{74.13}$_{\pm0.91}$ & \textbf{53.10}$_{\pm0.13}$ & \textbf{0.68}$_{\pm0.01}$ & \textbf{316.47}$_{\pm1.46}$ & \underline{0.00}$_{\pm0.00}$ \\
\cmidrule{2-11}
 & \multirow{2}{*}{JAMIE} & fewshot & \underline{71.10}$_{\pm1.03}$ & \textbf{66.41}$_{\pm4.08}$ & \underline{70.89}$_{\pm1.04}$ & \underline{72.23}$_{\pm0.73}$ & \underline{11.07}$_{\pm0.11}$ & \underline{1.03}$_{\pm0.01}$ & \underline{8.37}$_{\pm0.72}$ & \textbf{22.93}$_{\pm3.00}$ \\
 &  & normal & \textbf{71.80}$_{\pm1.05}$ & \underline{53.51}$_{\pm5.40}$ & \textbf{70.93}$_{\pm1.09}$ & \textbf{75.68}$_{\pm0.66}$ & \textbf{14.60}$_{\pm0.14}$ & \textbf{0.98}$_{\pm0.01}$ & \textbf{6.28}$_{\pm0.20}$ & \underline{70.02}$_{\pm1.45}$ \\
\cmidrule{2-11}
 & \multirow{2}{*}{scButterfly} & fewshot & \underline{68.56}$_{\pm0.73}$ & \underline{28.27}$_{\pm2.65}$ & \underline{68.13}$_{\pm0.76}$ & \underline{84.23}$_{\pm0.79}$ & \textbf{30.04}$_{\pm0.06}$ & \textbf{0.00}$_{\pm0.00}$ & \textbf{26.49}$_{\pm1.68}$ & \textbf{0.00}$_{\pm0.00}$ \\
 &  & normal & \textbf{74.15}$_{\pm3.15}$ & \textbf{41.93}$_{\pm2.65}$ & \textbf{73.09}$_{\pm3.17}$ & \textbf{84.64}$_{\pm5.37}$ & \underline{28.41}$_{\pm5.47}$ & \underline{0.00}$_{\pm0.00}$ & \underline{152.59}$_{\pm329.95}$ & \underline{27.91}$_{\pm15.92}$ \\
\cmidrule{2-11}
 & \multirow{2}{*}{multiDGD} & fewshot & \underline{48.59}$_{\pm1.48}$ & \underline{12.40}$_{\pm0.88}$ & \underline{47.71}$_{\pm1.50}$ & \underline{62.60}$_{\pm2.07}$ & \underline{52.58}$_{\pm0.89}$ & \textbf{0.81}$_{\pm0.00}$ & \underline{433.06}$_{\pm0.51}$ & \textbf{0.00}$_{\pm0.00}$ \\
 &  & normal & \textbf{53.49}$_{\pm1.01}$ & \textbf{19.56}$_{\pm1.56}$ & \textbf{51.33}$_{\pm1.13}$ & \textbf{63.19}$_{\pm0.64}$ & \textbf{54.04}$_{\pm1.35}$ & \underline{0.81}$_{\pm0.01}$ & \textbf{424.18}$_{\pm1.71}$ & \underline{0.00}$_{\pm0.00}$ \\
\cmidrule{2-11}
 & \multirow{2}{*}{scDiffusionX} & fewshot & \textbf{62.85}$_{\pm2.98}$ & \textbf{48.02}$_{\pm8.41}$ & \textbf{62.57}$_{\pm3.01}$ & \textbf{64.58}$_{\pm2.42}$ & \textbf{44.08}$_{\pm0.14}$ & \underline{0.12}$_{\pm0.00}$ & \textbf{177.53}$_{\pm9.98}$ & \textbf{0.00}$_{\pm0.00}$ \\
 &  & normal & \underline{37.64}$_{\pm2.27}$ & \underline{18.51}$_{\pm1.95}$ & \underline{35.80}$_{\pm2.35}$ & \underline{39.44}$_{\pm2.75}$ & \underline{43.09}$_{\pm0.26}$ & \textbf{0.12}$_{\pm0.00}$ & \underline{190.85}$_{\pm5.69}$ & \underline{0.00}$_{\pm0.00}$ \\
\bottomrule
\end{tabular}
\end{table*}

\end{document}